%% file: acl_latex.tex
\documentclass[11pt]{article}

\usepackage[final]{acl}

\usepackage{xcolor}

\usepackage{times}
\usepackage{latexsym}

\usepackage[T1]{fontenc}
\usepackage[utf8]{inputenc}

\usepackage{microtype}

\usepackage{inconsolata}

\usepackage{graphicx}
\usepackage{preamble}

\title{Grounding Agent Memory in Contextual Intent}

\author{
  \textbf{Ruozhen Yang\textsuperscript{1*}}\quad
  \textbf{Yucheng Jiang\textsuperscript{2*}}\quad
  \textbf{Yueqi Jiang\textsuperscript{1}}\\
  \textbf{Priyanka Kargupta\textsuperscript{1}}\quad
  \textbf{Yunyi Zhang\textsuperscript{1}}\quad
  \textbf{Jiawei Han\textsuperscript{1}}
  \\
  \textsuperscript{1}University of Illinois Urbana-Champaign\quad
  \textsuperscript{2}Stanford University\\
  \texttt{\{ruozhen2, hanj\}@illinois.edu}\quad \texttt{yuchengj@cs.stanford.edu}
}

\begin{document}
\maketitle
\input{sections/0_abstract}

\etocdepthtag.toc{mtchapter}
\section{Introduction}
\label{sec:introduction}
\input{sections/1_introduction}

\section{Methods}
\label{sec:contextual_intent}
\input{sections/2_contextual_intent}

\section{\benchmarkname}
\label{sec:benchmark}
\input{sections/3_castle_bench}

\section{Experiments}
\label{sec:experiments}
\input{sections/4_experiments}

\section{Results and Analysis}
\label{sec:results}
\input{sections/5_results}

\section{Related Works}
\label{sec:related}
\input{sections/6_related_works}

\section{Conclusion}
\label{sec:conclusion}
\input{sections/7_conclusion}

\section*{Limitations}

\input{sections/8_limitation}

\section*{Ethical Considerations}
\input{sections/9_ethics}

\section*{Acknowledgements}
\input{sections/99_acknowledgement}

\bibliography{custom}

\clearpage
\appendix
\label{sec:appendix}
\input{sections/999_appendix}

\end{document}

%% file: sections/0_abstract.tex
\begin{abstract}
Deploying large language models in long-horizon, goal-oriented interactions remains challenging because similar entities and facts recur under different latent goals and constraints, causing memory systems to retrieve context-mismatched evidence. We propose \textbf{\methodname{}} (\textbf{S}tructured \textbf{I}ntent \textbf{T}racking in \textbf{C}ontextual \textbf{H}istory), an agentic memory system that \emph{indexes} each trajectory step with a structured retrieval cue, \textit{contextual intent}, and retrieves history by matching the current step’s intent. Contextual intent provides compact signals that disambiguate repeated mentions and reduce interference: (1) the current latent goal defining a thematic segment, (2) the action type, and (3) the salient entity types anchoring which attributes matter. During inference, \methodname{} filters and prioritizes memory snippets by intent compatibility, suppressing semantically similar but context-incompatible history.

For evaluation, we introduce \textbf{\benchmarkname{}}, a benchmark for context-aware retrieval in realistic, dynamic, goal-oriented trajectories. Across \benchmarkname{} and LongMemEval, \methodname{} achieves state-of-the-art performance, outperforming the strongest baseline by 35.6\%, with the largest gains as trajectory length increases. Our analysis shows that intent indexing substantially reduces retrieval noise, supporting intent-aware memory for robust long-horizon reasoning.\footnote{The codebase and benchmark are available at \url{https://contextual-intent.github.io/}.}

\end{abstract}

%% file: sections/1_introduction.tex
Large language models (LLMs) are deployed in \emph{long-horizon} tasks \cite{lost-in-middle} that require agents to track interleaved goals, resolve references to prior information, and coordinate actions over extended trajectories. Such settings arise across diverse applications, including extended human--agent dialogues \citep{costorm}, deep research workflows \citep{storm}, and autonomous tool-augmented environments \citep{OSWorld}.

\begin{figure}[t]
    \centering
    \includegraphics[width=\linewidth]{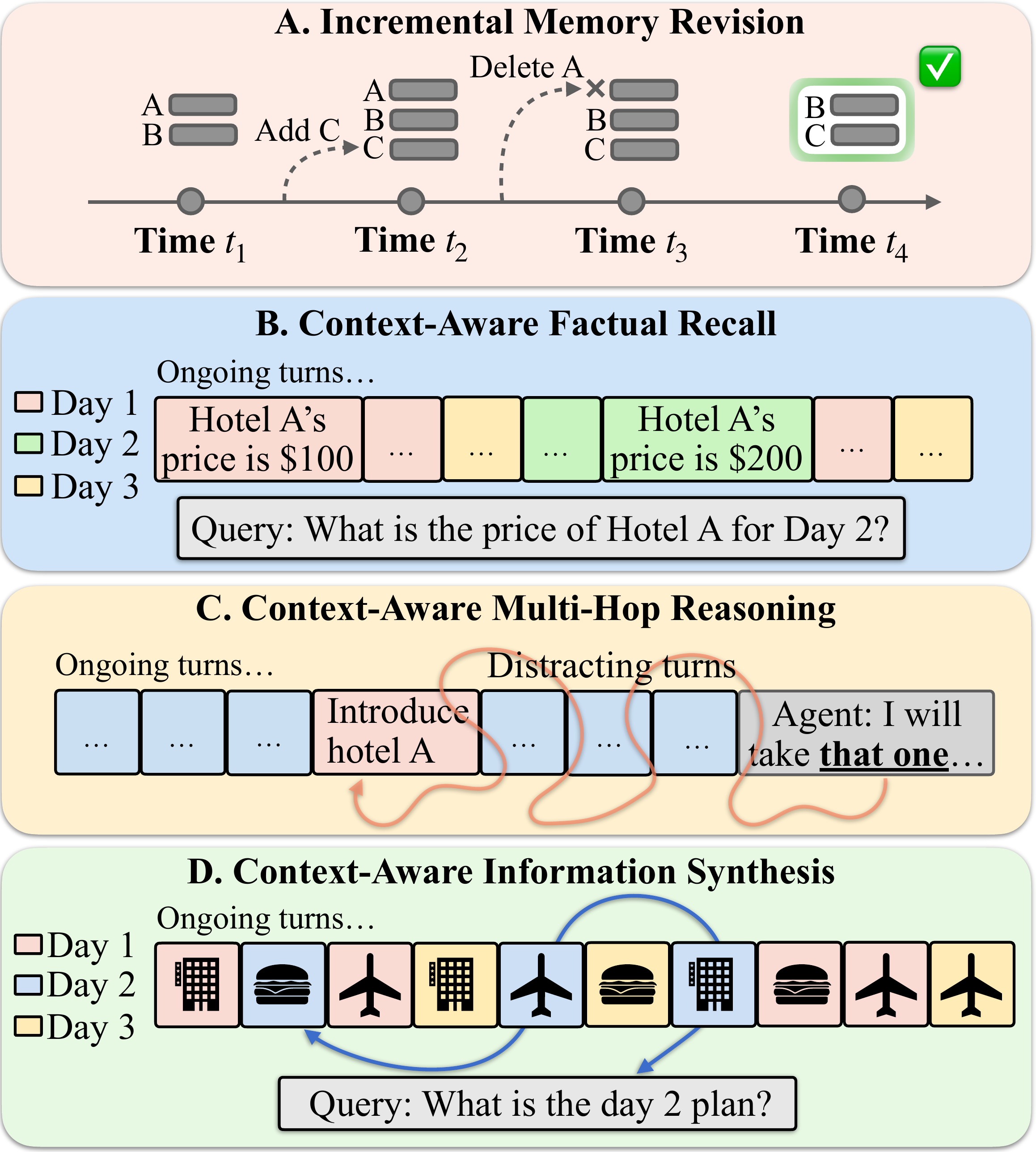}
    \caption{\textbf{The Challenges of Long-Horizon Agentic Memory.} We identify four capabilities required for robust agentic memory: (A) Incremental Memory Revision (tracking state changes over time); (B) Context-Aware Factual Recall (distinguishing semantically similar facts by context); (C) Context-Aware Multi-Hop Reasoning (resolving implicit references across distracting turns); and (D) Context-Aware Information Synthesis.}
    \label{fig:teaser}
    \vspace{-15pt}
\end{figure}

To address the complexity of these environments, recent works (e.g., \citealp{secom, amem, graphrag}) have introduced \emph{agentic memory} mechanisms that organize context online and \emph{retrieve} relevant information at inference time. This retrieval-centric view aligns with the notion of \emph{retrieval cues} in cognitive science \citep{tulving1973encoding, craik1972levels}: structured indices that determine which stored context information is accessed. In long-horizon settings, where many steps are semantically similar but contextually distinct, retrieval is bottlenecked by cue quality. Effective agentic memory must construct cues that reliably surface the right context amidst noisy, interference-prone history.

However, existing agentic memory systems remain misaligned with the demands of long-horizon, goal-oriented reasoning in Figure~\ref{fig:teaser}. Memory compression techniques \cite{raptor, secom} summarize semantically or temporally adjacent segments but often fail to link \emph{interleaved, persistent} goal contexts across non-adjacent segments (\reffig{fig:teaser}(C)), complicating state tracing (\reffig{fig:teaser}(A)) and information synthesis (\reffig{fig:teaser}(D)). Similarly, knowledge-graph-based memories \citep{graphrag, amem} encode entity-level relations but often lack explicit goal- or episode-level disambiguation, so repeated mentions of the same entities can correspond to distinct latent goals (\reffig{fig:teaser}(B)). While long-context LLMs \citep{gpt41, gemini25} can attend to extended trajectories, they become ineffective once trajectories exceed context limits and incur substantial overhead for real-time retrieval.

To address these limitations, we propose \textbf{\methodname{}} (\refsec{sec:contextual_intent}), an agentic memory system that models the latent intent underlying each step of a task trajectory. Grounded in Event Structure Theory \citep{zacks2001event}, \methodname{} is motivated by the hypothesis that robust recall requires organizing experience by (i) the superordinate goal context and (ii) recurring action categories. We instantiate \textbf{contextual intent}, a structured retrieval cue induced online without a fixed ontology:
(1) \textbf{thematic scope} (\S~\ref{sec:thematic_scope}), a stable segment label tracking the current goal (e.g., \textit{``Day 2 Itinerary''} or \textit{``Model Optimization''});
(2) \textbf{event type} (\S~\ref{sec:event_label}), an action label capturing the operation performed (e.g., \textit{``Rebuttal''}, \textit{``Hyperparameter Tuning''}, \textit{``Price Inquiry''}); and
(3) \textbf{key entity types} (\S~\ref{sec:entity_label}), the entity classes that determine which attributes are relevant in-context (e.g., \textit{Metric}, \textit{Hyperparameter}, \textit{Price}, \textit{Rating}).
Together, these cues explicate contextual information otherwise ambiguous in the raw trajectory: thematic scope links non-adjacent, goal-consistent steps for multi-hop reasoning, state tracing, and synthesis (\reffig{fig:teaser}(C, A, D)), while event and entity typing disambiguate repeated mentions by anchoring facts to functional roles and expected attributes, improving recall under interference (\reffig{fig:teaser}(B)). At inference time, \methodname{} filters and prioritizes memory snippets by structural compatibility, suppressing semantically similar but context-mismatched content (\S~\ref{sec:retrieval}).

A parallel gap lies in how we evaluate long-horizon memory: existing benchmarks rarely test whether models can retrieve the \emph{right} fact \emph{in the right context}. In realistic agentic settings, success is not determined solely by finding semantically relevant text, but by verifying that retrieved information matches the contextual constraints under which it was stated or requested (e.g., a hotel ``price'' is only meaningful when conditioned on dates, location, and room type). Yet current evaluations often sidestep this requirement. First, many benchmarks segment a long trajectory into largely independent mini-episodes or topic blocks \citep{longmemeval}, which makes retrieval effectively local and reduces the need to track global state across interleaved goals. Second, they enforce strict turn-taking between a user and an assistant \citep{longmemeval, locomo}, where queries are typically answered immediately after they are posed; this adjacency masks the harder case where requests are interleaved, deferred, and resolved only after several intervening steps.

To address these evaluation gaps, we introduce \textbf{\benchmarkname{}} (Context-Aware Agent Memory Evaluation Benchmark) (\refsec{sec:benchmark}), which targets \emph{context-aware retrieval} in long, goal-oriented trajectories. \benchmarkname{} stresses this capability along three axes: interleaved, non-turn-taking interaction structure; multi-domain coverage; and controlled difficulty via diverse question types and length-stratified subsets. These choices make dispersed, often implicit constraints necessary for answering and reveal a complementary failure mode obscured in existing evaluations where strong baselines are near-saturated.

Our contributions are threefold:
\begin{enumerate}
    \setlength{\itemsep}{4pt}
    \setlength{\parskip}{0pt}
    \item We propose \methodname{}, an intent-aware, domain-agnostic agentic memory system that induces contextual intent online and uses intent compatibility to filter and prioritize retrieved history.
    \item We introduce \benchmarkname{}, a multi-domain benchmark for \emph{context-aware retrieval} in long, goal-oriented trajectories, with length-stratified subsets.
    \item Empirically, \methodname{} consistently outperforms strong long-context and structured-memory baselines on \benchmarkname{}, validating the importance of intent-aware retrieval for robust long-horizon reasoning.
\end{enumerate}

%% file: sections/2_contextual_intent.tex
\subsection{Contextual Intent Schema}
\label{subsec:schema}

Our approach, \textbf{ST}ructured \textbf{I}ntent \textbf{T}racking in \textbf{C}ontextual \textbf{H}istory (\methodname), is grounded in Event Structure Theory \citep{zacks2001event}, which argues that humans comprehend extended activity by building event representations that support prediction and memory. A key implication is that extended experience is not treated as a single flat stream: people naturally organize behavior into segments that maintain a coherent goal context, while also recognizing recurring kinds of actions that reappear across contexts. We mirror this organization with three complementary cues induced online from the trajectory (i.e., without assuming a domain-specific ontology), making the representation applicable across task domains. First, a \emph{partonomy} assigns each trajectory step to a \emph{thematic scope}---a segment label that summarizes the current goal context and remains stable until the latent goal shifts (e.g., \textit{``Day 2 Itinerary''}) (\refsec{sec:thematic_scope}). Second, a \emph{taxonomy} assigns each step an \emph{event type} that captures its basic operation (e.g., \textit{``searching''}, \textit{``comparing''}, \textit{``booking''}) regardless of which segment it occurs in; the label set is derived on the fly and can expand as new operations appear (\refsec{sec:event_label}). Third, to support retrieval of factual details, we extract \emph{key entity types} that identify the classes of information necessary to satisfy the current goal \citep{talmy1975semantics} (e.g., {\textit{Price}, \textit{Rating}}), providing a lightweight schema template that anchors which details should be surfaced to make progress on the task (\refsec{sec:entity_label}). \methodname{} tracks these signals at each step and uses them as retrieval cues to retrieve the correct context at inference time.

We formalize these cues as \textbf{\emph{Contextual Intent}}, a structured representation of the latent goal used as a retrieval cue. For a task trajectory $T = \{s_1, s_2, \ldots, s_n\}$ where each step is a tuple $s_t = (r_t, a_t, \tau_t)$ specifying the acting agent role $r_t$, a natural-language description of the action $a_t$, and a timestamp $\tau_t$, we infer a contextual intent tuple $
\iota_t = (\sigma_t, \epsilon_t, \kappa_t)$,
where $\sigma_t$ denotes the thematic scope, $\epsilon_t$ is the event type, and $\kappa_t$ represents the set of key entity types.

\subsection{Contextual Intent Construction}
\label{subsec:construction}

\begin{figure*}[t]
    \centering
    \includegraphics[width=\textwidth]{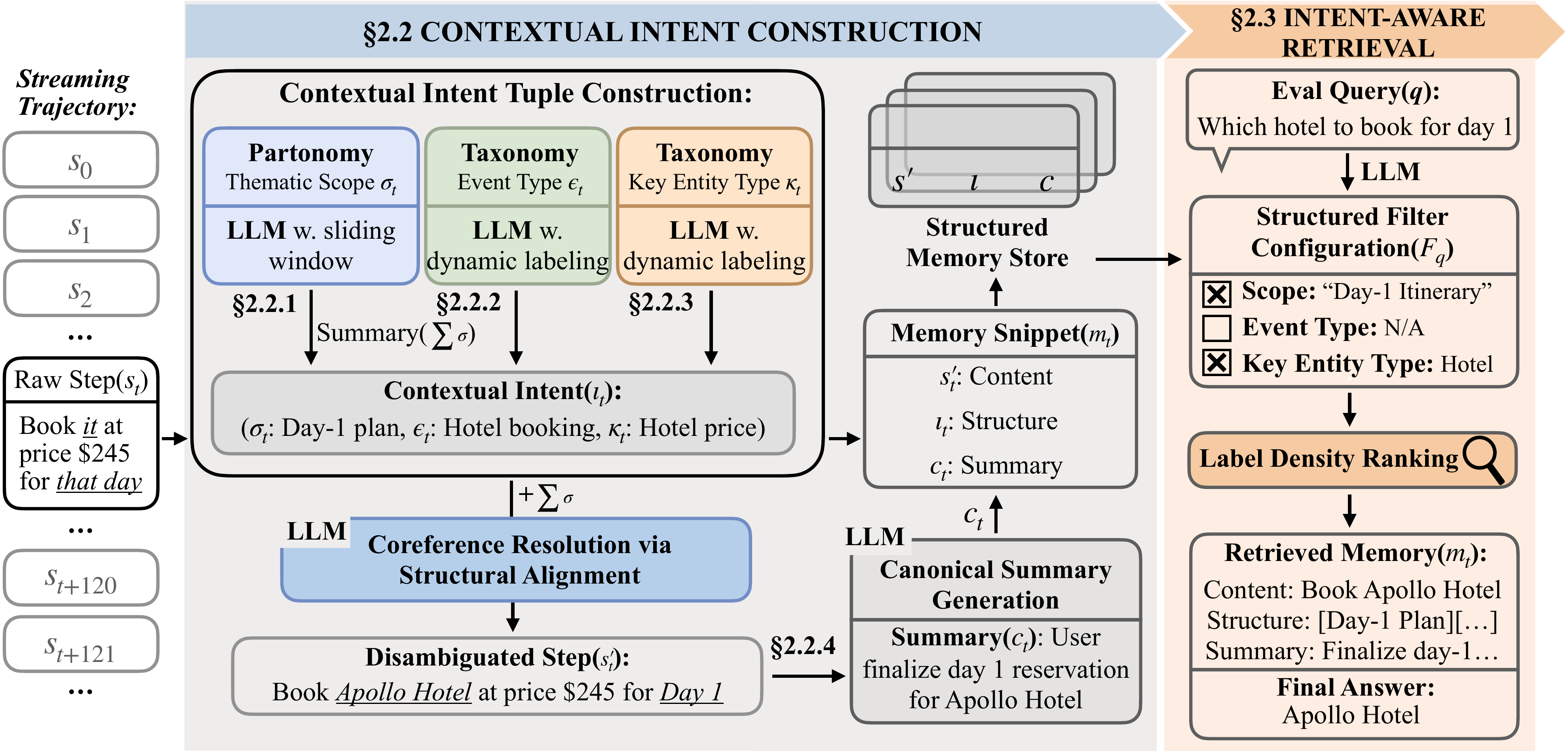}
    \caption{\textbf{Overview of \methodname.} The framework operates in two phases. \textbf{Left (\S\ref{subsec:construction}): Contextual Intent Construction.} From a streaming trajectory, the model dynamically induces three structural cues—Thematic Scope ($\sigma_t$), Event Type ($\epsilon_t$), and Key Entity Types ($\kappa_t$)—to form a \emph{Contextual Intent} tuple $\iota_t$. This structure guides coreference resolution (e.g., resolving ``it'') and summary generation to create a structured Memory Snippet. \textbf{Right (\S\ref{sec:retrieval}): Intent-Aware Retrieval.} During inference, the evaluation query $q$ is mapped to a structured filter configuration $F_q$. The retrieval engine applies \emph{Label Density Ranking}, prioritizing memories that strictly match the intended structure before ranking by semantic content.}
    \label{fig:method}
\end{figure*}

\subsubsection{Thematic Scope Induction.}
\label{sec:thematic_scope}
Event Perception literature suggests that humans naturally segment continuous activity into discrete units at points of maximal feature change, which correlate strongly with sub-goal completion \citep{newtson1977objective}. While this segmentation captures local intent, treating segments in isolation leads to context loss, since coarse-grained goals (the ``macro-structure'') often persist across fine-grained boundaries. To model this \emph{partonomy}, we induce a \textbf{thematic scope} $\sigma_t$ that acts as a coarse-grained container for fine-grained steps. Concretely, we use a sliding window in which an LLM predictor $\mathcal{M}$ analyzes the current step $s_t$ with a recent history buffer $H_{\text{scope}}$ and the previously induced scope to detect shifts in the latent goal: $\sigma_t = \mathcal{M}_{\text{scope}}(s_t, H_{\text{scope}}, \sigma_{t-1})$.
Consistent with findings that ``behavior episodes'' \citep{barker1954midwest} are characterized by continuity in setting and mental sphere \citep{zacks2001event}, $\sigma_t$ remains constant until the model detects a boundary event (i.e., a goal-state divergence). When a boundary is detected by the LLM, a new scope label is induced; otherwise, the step inherits the prior scope (e.g., persisting in ``\textit{Day 2 Itinerary}'' for travel, or ``\textit{Model Optimization}'' in machine learning). To prevent context overflow while preserving the ongoing episode's ``gist'', we maintain a compressed summary $\Sigma_{\sigma}$ of the current scope online, which is then provided to subsequent scope predictions (see Appendix \ref{app:stitch-prompts} for prompts).

\subsubsection{Taxonomic Event Labeling.}
\label{sec:event_label}
While thematic scope captures the \emph{episodic} structure of a specific activity, \textbf{event labeling} captures \emph{taxonomic} regularities that recur across distinct scopes. We induce event types directly from the trajectory without a pre-defined ontology, mirroring how humans learn event structure via statistical covariation \citep{avrahami1994emergence}; this allows the taxonomy to naturally adapt to diverse task domains (see Appendix \ref{app:stitch-prompts} for prompts).

We adopt a dynamic label-space evolution strategy. First, we generate a seed vocabulary of event types $\mathcal{V}_{\epsilon}$ via zero-shot inference on the initial $N_{\text{start}}$ trajectory steps, producing task-appropriate, interpretable labels. During online processing, for each step $s_t$, we retrieve the top-$k_{\text{event}}$ semantically similar labels from $\mathcal{V}_{\epsilon}$ and prompt $\mathcal{M}_{\text{label}}$ to select the best-fitting event label $\epsilon_t = \mathcal{M}_{\text{label}}\!\bigl(s_t, \text{Retrieve}(\mathcal{V}_{\epsilon}, s_t, k_{\text{event}})\bigr)$.
If no suitable label is found, we introduce a new label into $\mathcal{V}_{\epsilon}$. To maintain a compact and distinct taxonomy, we perform periodic refinement every $k_{\text{update}}$ steps to consolidate semantically overlapping (near-synonymous) labels.
%Analogous to how objects have a cognitively accessible ``basic level'' of categorization (e.g., \textit{``chair''} vs.\ \textit{``furniture''}) \citep{rosch1978principles}, actions admit basic-level categories that abstract away particulars (e.g., \textit{Rebuttal''} in debate, \textit{Hyperparameter Tuning''} in ML research, or \textit{Price Inquiry''} in travel). 

\subsubsection{Key Entity Type Extraction.}
\label{sec:entity_label}
Events are often defined by interactions between an actor and specific objects or ``figures'' \citep{talmy1975semantics}. To anchor retrieval of factual details, we extract \textbf{key entity types} $\kappa_t$, representing the \emph{classes} of detailed information required (e.g., distinguishing \textit{``Metric''} from \textit{``Hyperparameter''} in research; distinguishing \textit{``Price''} from \textit{``Rating''} in travel). This yields a lightweight ``schema template'' that enhances generalization, enabling retrieval to target the specific entity \emph{types} required by the current intent rather than being misled by ambiguous surface forms with functionally distinct roles. As with event labeling, entity types are induced online without assuming a fixed domain schema, thus ensuring generalization (see Appendix \ref{app:stitch-prompts} for prompts).

We maintain a growing label space of entity types $\mathcal{V}_{\kappa}$. For each step, the model identifies necessary semantic entities and maps them to existing types in $\mathcal{V}_{\kappa}$ or generates new types if the class is novel $
\kappa_t = \mathcal{M}_{\text{entity}}(s_t, \mathcal{V}_{\kappa})$.
Similar to the event taxonomy, $\mathcal{V}_{\kappa}$ undergoes periodic consolidation to merge semantically overlapping entity classes.

\paragraph{Coreference Resolution via Structural Alignment.}
Extended goal-directed trajectories often rely on implicit references to prior states (e.g., \textit{``Book \textbf{it}.''}). \citet{zacks2001event} note that event coherence relies on tracking participants across temporal boundaries. We resolve these references by leveraging the induced partonomic and taxonomic structure. When a step $s_t$ contains ambiguous references, we retrieve a context set $C_{\text{align}}$ consisting of prior steps that share the same 
\emph{thematic scope} $\sigma_t$ and/or compatible \emph{event types} $\epsilon_t$. This structurally aligned context is then used to rewrite $s_t$ into a disambiguated form $s'_t = \mathcal{M}_{\text{rewrite}}(s_t, C_{\text{align}})$.
This ensures that pronouns (e.g., \textit{``it''}) are resolved to their referents (e.g., \textit{``the Apollo Hotel''}) prior to storage. We validate the efficacy of this rewriting module via an automated audit of \emph{Entity Resolution Recall}, ensuring that implicit references are correctly grounded to their canonical entity names (see Appendix~\ref{app:stitch-prompts} for prompt and ~\ref{app:coref_verification} for verification details).

\subsubsection{Memory Snippet Construction}
\label{sec:memory_snippet}
To support downstream inference, we augment the raw step content with its inferred structural properties. We generate a canonical summary $c_t$---a concise natural-language description encapsulating the agentic operation and its contextual intent---via $
c_t = \mathcal{M}_{\text{sum}}(s'_t, \iota_t)$.
The final \textbf{Memory Snippet} is defined as $m_t = (s_t, \iota_t, c_t)$. This composite representation enables retrieval to query the \emph{structure} of the past (via $\iota_t$) alongside the \emph{content} (via $c_t$).

\subsection{Intent-Aware Retrieval}
\label{sec:retrieval}
During inference, rather than relying solely on unstructured semantic similarity, we perform \emph{intent-aware} filtering using contextual intent. Given a query $q$, we prompt an LLM to produce a \emph{filter configuration} $F_q$ that specifies target values over the same schema components as $\iota_t$, i.e.,
$F_q = (\mathcal{S}_q, \mathcal{E}_q, \mathcal{K}_q)$,
where $\mathcal{S}_q$ is a set of candidate thematic scopes, $\mathcal{E}_q$ a set of candidate event types, and $\mathcal{K}_q$ a set of key entity types. In practice, the LLM is conditioned on $q$ and the current label inventories (i.e., the induced scope labels and the vocabularies $\mathcal{V}_{\epsilon}$ and $\mathcal{V}_{\kappa}$) to ensure that $F_q$ is expressed in the same label space as stored memories. (Prompt templates and examples are provided in Appendix~\ref{app:stitch-prompts}.)
This mirrors how humans recall experiences by indexing through a goal episode (partonomy) and/or a recurring operation (taxonomy), while using expected participants to anchor required factual details.

We then rank memory snippets via \emph{structural alignment} with $F_q$, utilizing semantic similarity as a secondary tie-breaker. We employ a \emph{label density ranking} strategy, which prioritizes snippets based on the cardinality of their overlap with the filter configuration (i.e., the number of satisfied constraints in $\iota_t$). This strategy serves as a proxy for contextual relevance: it ensures that memory snippets sharing the highest degree of \emph{contextual intent} with the query---and therefore possessing the strongest contextual affinity to the user's information need---are strictly prioritized. Among snippets with equivalent label density, we rank by semantic similarity between the query and content (e.g., $\mathrm{sim}(q, c_t)$). Finally, we return the top-$k_{\text{retrieve}}$ memory snippets.

%% file: sections/3_castle_bench.tex
We introduce the \textbf{Context-aware Agent Memory Evaluation Benchmark} (\benchmarkname), a synthetic dataset designed to evaluate agentic memory in continuous, goal-oriented trajectories characterized by high \emph{contextual interference}. \benchmarkname\ constructs dense interaction histories where recurring entities and interleaved goals create significant ambiguity, specifically designed to evaluate an agent's ability to manage context over long horizons.

Formally, each benchmark instance consists of a trajectory $T = \{s_1, s_2, \ldots, s_n\}$, where each step is defined as $s_t = (r_t, a_t, \tau_t)$. During evaluation, models are presented with a set of free-response questions $Q$ and must generate answers by reasoning over the full trajectory. To assess robust generalization, \benchmarkname\ comprises trajectories spanning distinct domains.

We construct \benchmarkname\ according to three design principles: (i)~\emph{symbolically grounded causal consistency}, where we decouple storyboard planning from surface generation to ensure logically coherent long-horizon state transitions; (ii)~\emph{controlled semantic interference}, utilizing a closed-world environment to densely reuse static entities and challenge models to disambiguate fine-grained context; and (iii)~\emph{dynamic referential ambiguity}, which necessitates resolving indirect pragmatic references and tracking incremental state updates rather than relying on static retrieval.

Based on these principles, we instantiate the benchmark in two domains: \textbf{(i)~Travel Planning}, in which an LLM-simulated assistant collaborates with a simulated user to construct multi-day itineraries; and \textbf{(ii)~Debate}, in which two LLM agents advocate opposing positions in evidence-grounded policy argumentation.

\subsection{Benchmark Curation}
\label{sec:benchmark_curation}

Inspired by prior work showing that intermediate structured representations improve dataset consistency \citep{rastogi2020towards, eric-etal-2020-multiwoz}, we design a four-stage pipeline that decouples symbolic planning from natural-language generation. This approach allows us to enforce complex causal dependencies and maintain logically coherent state transitions prior to surface realization.

As detailed in Appendix~\ref{app:curation_pipeline} and \ref{app:domain_specs}, our pipeline consists of: (1) closed-world environment construction, creating a controlled set of static entities and knowledge to mitigate contamination; (2) symbolic storyboard planning, where we predefine action spaces and generate symbolic operations (e.g., latent goals, payloads) subject to pragmatic constraints; (3) storyboard-conditional trajectory generation, translating symbolic plans into natural language using LLMs; and (4) pragmatic refinement, applying referential remodeling and turn segmentation to mimic natural communication patterns and increase difficulty. This structured generation enables faithful automatic annotation of evaluation questions.

\subsection{Evaluation Questions and Metrics}
\label{sec:benchmark_question_types}

For each trajectory, we pose four categories of free-response questions that target complementary aspects of context-aware retrieval and long-horizon reasoning (Figure~\ref{fig:teaser}). See Appendix~\ref{app:benchmark_example} for illustrative examples; Appendices~\ref{app:appendix_ground_truth} and \ref{app:qa} provide details on ground-truth answer generation and human verification.

\vspace{5pt}
\noindent\textbf{Type 1: Incremental Memory Revision.} Following \cite{longmemeval}, these questions assess the ability to maintain and update entity states as they evolve (e.g., tracking restaurant candidates as they are added or rejected across turns).

\vspace{5pt}
\noindent\textbf{Type 2: Context-Aware Factual Recall.} These questions test precise information retrieval amidst semantically similar content. Models must disambiguate details that differ only by context (e.g., retrieving a hotel price for Day~1 vs. Day~2).

\vspace{5pt}
\noindent\textbf{Type 3: Context-Aware Multi-Hop Reasoning.} These questions require resolving referential expressions before retrieval. For example, answering a question about ``the dinner reservation I mentioned earlier'' requires first identifying the referent, then retrieving associated attributes.

\vspace{5pt}
\noindent\textbf{Type 4: Information Synthesis.} These questions assess the ability to aggregate information distributed across multiple steps, such as reconstructing a complete itinerary from scattered bookings and visits.

\paragraph{Metrics.}
We employ answer-set macro-averaged precision, recall, and F1 score. This metric provides a more nuanced assessment for multi-element answers and captures partial credit scenarios better than binary accuracy.

%% file: sections/4_experiments.tex
\input{tables/main_table}
\paragraph{Benchmarks.}
We evaluate \methodname{} on three long-context QA benchmarks: (i) \benchmarkname\ (\S\ref{sec:benchmark}), our multi-agent trajectory benchmark; question types and metrics are defined in \S\ref{sec:benchmark_question_types}; (ii) LongMemEval \citep{longmemeval}, which embeds task-relevant sessions into unrelated chat history and is stratified by context length,\footnote{For computational efficiency, we randomly subsample each LongMemEval subset from the original 500 questions.} making it the closest public benchmark to our setting; and (iii) LoCoMo \citep{locomo}, which evaluates conversational memory in dialogue agents,\footnote{For a fair comparison with text-only baselines, we exclude the subset of LoCoMo questions that require visual understanding.} serving as a broader transfer benchmark beyond persistent task-state tracking. Following prior work, we report accuracy on LongMemEval and LoCoMo.

\paragraph{Baselines} We compare our method against 13 baselines categorized into three groups: 
\begin{itemize} 
    \setlength{\itemsep}{4pt}
    \setlength{\parskip}{0pt}
    \item \textbf{Long-Context LLMs (Zero-Shot):} We evaluate powerful models operating directly on the full context, including gpt-4o-mini, gpt-4.1-mini, gpt-5-mini, qwen3-235B-A22B-Instruct, and deepseek-v3.1. For inputs exceeding the context window, we apply left-truncation (keeping the most recent history). 
    \item \textbf{RAG Baselines:} We evaluate standard Retrieval-Augmented Generation using dense retrievers with three embedding models: qwen3-embedding-8B, text-embedding-3-small, and text-embedding-3-large. 
    \item \textbf{Structured Memory Agents:} We compare against state-of-the-art memory-augmented systems: RAPTOR \citep{raptor}, GraphRAG \citep{graphrag}, HippoRAG 2 \citep{hipporag}, A-mem \citep{amem}, and Secom \citep{secom}. See detailed baseline descriptions in Appendix \ref{app:baseline_description}. 
\end{itemize}

\paragraph{Implementation details.}
All baseline methods and all intent construction stages of \methodname{} (\S\ref{subsec:construction}) use gpt-5-mini with medium reasoning effort as the backbone LLM generator; we set $N_{\text{start}}{=}50$, $k_{\text{update}}{=}50$, $k_{\text{retrieve}}{=}40$, and $k_{\text{event}}{=}5$.

To ensure fair comparison, we enforce a shared retrieval budget for all retrieval-based and memory-augmented methods: the inference context is capped at 4{,}096 tokens. If retrieved content exceeds the budget, we tail-truncate to the last complete sentence. We automatically evaluate all generated answers against ground truth using an LLM-as-a-judge with gpt-4.1-mini (\texttt{temperature=0.0}, \texttt{top\_p=0.9}); we use identical generation and judging prompts across methods (Appendix~\ref{app:prompt_answer_gen_eval}).

%% file: tables/main_table.tex
\begin{table*}[t]
\centering
% \small % Font size is now handled by resizebox scaling
\setlength{\tabcolsep}{3pt} 
\resizebox{\textwidth}{!}{% 
\begin{tabular}{lcccccccccc} % Adjusted column count to 11
\toprule
\multirow{2}{*}{\textbf{Model}} & \multicolumn{6}{c}{\textbf{\benchmarkname}} & \multicolumn{3}{c}{\textbf{LongMemEval}} & \multicolumn{1}{c}{\textbf{LoCoMo}} \\ 
\cmidrule(lr){2-7} \cmidrule(lr){8-10} \cmidrule(lr){11-11}
 & $\text{Prec}_{S}$ & $\text{F1}_{S}$ & $\text{Prec}_{M}$ & $\text{F1}_{M}$ & $\text{Prec}_{L}$ & $\text{F1}_{L}$ & $\text{Acc}_{O}$ & $\text{Acc}_{S}$ & $\text{Acc}_{M}$ & $\text{Acc}$ \\
\midrule
\multicolumn{11}{l}{\textit{Long Context Agents}} \\
DeepSeek V3.1 (128k) & 0.238 & 0.228 & 0.012 & 0.010 & 0.000 & 0.000 & 0.620 & 0.240 & 0.267 & 0.587\\
Qwen3 (256k) & 0.724 & 0.704 & 0.192 & 0.152 & 0.016 & 0.016 & 0.740 & 0.160 & 0.333 & 0.693\\
GPT-4o-mini (128k) & 0.336 & 0.315 & 0.082 & 0.075 & 0.017 & 0.017 & 0.560 & 0.060 & 0.267 & 0.571\\
GPT-4.1-mini (1M) & 0.715 & 0.712 & 0.401 & 0.362 & 0.317 & 0.213 & 0.720 & 0.200 & 0.067 & 0.682\\
GPT-5-mini (400k) & 0.775 & 0.804 & 0.576 & 0.566 & 0.268 & 0.212 & \textbf{0.860} & 0.820 & 0.533 & \textbf{0.811}\\

\midrule
\multicolumn{11}{l}{\textit{Embedding RAG Agents}} \\
Qwen 8B & 0.306 & 0.280 & 0.176 & 0.157 & 0.172 & 0.152 & 0.740 & 0.780 & 0.533 & 0.511\\
text-embedding-3-small & 0.275 & 0.269 & 0.224 & 0.195 & 0.219 & 0.182 & 0.800 & 0.800 & 0.200 & 0.593\\
text-embedding-3-large & 0.346 & 0.317 & 0.190 & 0.168 & 0.236 & 0.195 & 0.800 & 0.800 & 0.267 & 0.661\\
\midrule
\multicolumn{11}{l}{\textit{Structure-Augmented RAG Agents}} \\
RAPTOR \citep{raptor} & 0.343 & 0.329 & 0.122 & 0.117 & 0.168 & 0.139 & 0.680 & 0.480 & 0.467 & 0.671\\
GraphRAG \citep{graphrag} & 0.385 & 0.371 & 0.189 & 0.165 & 0.196 & 0.156 & 0.820 & 0.840 & 0.667 & 0.648\\
HippoRAG 2 \citep{hipporag} & 0.405 & 0.390 & 0.211 & 0.191 & 0.212 & 0.186 & 0.820 & 0.800 & 0.667 & 0.725\\
A-mem \citep{amem} & 0.387 & 0.376 & 0.217 & 0.196 & 0.216 & 0.186 & 0.780 & 0.740 & 0.667 & 0.731\\
Secom \citep{secom} & 0.532 & 0.501 & 0.141 & 0.114 & 0.282 & 0.236 & 0.520 & 0.580 & 0.600 & 0.640\\
\midrule
\methodname (Ours) & \textbf{0.810} & \textbf{0.844} & \textbf{0.665}$^{\dagger}$ & \textbf{0.682}$^{\dagger}$ & \textbf{0.616}$^{\dagger}$ & \textbf{0.592}$^{\dagger}$ & \textbf{0.860} & \textbf{0.860} & \textbf{0.800} & 0.703\\

\bottomrule
\end{tabular}%
}
\caption{Performance on \benchmarkname{}, LongMemEval, and LoCoMo. We report answer-set Macro Precision/Macro-F1 (omitting Recall for brevity) on \benchmarkname$_S$/$\!_M$/$\!_L$ ($N{=}144/168/61$), average Accuracy on LongMemEval$_O$/$\!_S$/$\!_M$ ($N{=}50/50/15$)~\citep{longmemeval}, and Accuracy on LoCoMo~\citep{locomo}. Entries marked `-' were not evaluated on LoCoMo. Following \citet{secom}, the memory-retrieval token budget is 4096. Best results are \textbf{bold}. $\dagger$: paired t-test vs. the strongest baseline within each \benchmarkname{} subset, $p{<}0.05$. Appendix~\ref{app:prompt_answer_gen_eval} details the prompts for answer generation and LLM-based evaluation.}

\label{tab:main_results}
\end{table*}

%% file: sections/5_results.tex
\paragraph{Overview.}
Table~\ref{tab:main_results} compares \methodname{} against baselines on \benchmarkname{}, LongMemEval, and LoCoMo. On \benchmarkname{}, \methodname{} achieves parity with the strongest baseline (gpt-5-mini) on the small subset while demonstrating substantially better scalability. As trajectory lengths increase by factors of $6$ and $17$ for the Medium and Large subsets, respectively, standard baselines degrade sharply. In contrast, \methodname{} remains robust, outperforming the best baseline by \textbf{11.6\%} (20.5\% relative) on the Medium subset and \textbf{35.6\%} (100\% relative) on the Large subset. On the public benchmarks, \methodname{} achieves the strongest results on LongMemEval and remains competitive on LoCoMo, consistent with its design for goal-oriented state tracking under contextual interference.

\paragraph{Analysis of Baselines.}
We observe distinct failure modes in the baseline methods. Long-context LLMs remain competitive on the small subset, effectively attending to scattered details. However, they suffer from substantial performance degradation on the Large subset, struggling to link multiple details across extended contexts (the ``lost-in-the-middle'' phenomenon). Conversely, structured agentic memory baselines perform adequately on simple factual recall but fail on distinguishing semantically similar facts that differ by their context (e.g., distinguishing a hotel price on Day 1 from Day 2). \methodname{} overcomes this by anchoring retrieval in the \emph{Contextual Intent} structure ($\iota_t$), ensuring that semantically similar information is correctly disambiguated by its thematic scope.

\subsection{Ablation Study}
\label{sec:ablation}
To understand the source of these gains, Table~\ref{tab:ablation_results} reports an ablation study over the components of contextual intent: thematic scope ($\sigma_t$), event type ($\epsilon_t$), key entity types ($\kappa_t$), and the coreference resolution module. We next analyze the dominant retrieval errors induced by question-time label selection; Appendix~\ref{app:error_analysis_label_overlap} provides additional diagnostics and methodology details.

\input{tables/ablation_table}

\paragraph{Thematic Scopes Effectively Reduce Context Noise.}
Thematic scope ($\sigma_t$) emerges as the most significant factor contributing to performance. By segmenting the trajectory into coarse-grained ``behavior episodes,'' $\sigma_t$ establishes necessary boundaries for memory retrieval. This proves critical for all question types, as it allows the model to constrain reasoning to relevant thematic segments, minimizing distraction from irrelevant history.

\paragraph{Granularity Trade-off.}
Taxonomic Event Labeling ($\epsilon_t$) improves performance on fine-grained lookup tasks but introduces instability in Question Type 4 (Information Synthesis). We attribute this to the inductive nature of our labeling process, which tends to generate highly specific event types (e.g., differentiating \textit{``Check Rates''} from \textit{``Book Hotel''}). While these fine-grained labels serve as effective filters for specific queries, they can hinder synthesis tasks that require aggregating information across multiple distinct action types. This suggests a trade-off: while a flat taxonomy is useful, a hierarchical structure is likely required to support both fine-grained filtering and coarse-grained integration.

\paragraph{Resolution Grounds Entities.}
The structural alignment coreference module serves as a foundational step. Its effectiveness depends heavily on the accuracy of the upstream thematic scope and event type assignments. By resolving ambiguous pronouns (e.g., \textit{``it''}) to their specific entity referents prior to storage, this module significantly boosts performance across all metrics by preventing the retrieval of ungrounded or ambiguous context snippets (see Appendix~\ref{app:coref_verification} for implementation details and results).

\subsection{Error Analysis}

To better understand the remaining failure modes of \methodname{}, we analyze the question-side label selection step in isolation. Across both domains, the selected labels almost never yield zero overlap with candidate turns, indicating that they provide meaningful constraints on the retrieval space. These results suggest that question-time labels function most effectively as a structural retrieval prior: they reliably concentrate retrieval on the relevant portion of the trajectory, while exact evidence selection emerges from their interaction with downstream ranking and reasoning.

We further examine the dominant causes of label-selection failure. As shown in Table~\ref{tab:label_failure_modes}, the most frequent error arises when the model selects labels that are plausible given the broader trajectory but are not supported by the question alone (78.4\%). The second most common failure is a mismatch in level of abstraction (71.8\%): the selected labels may be too coarse, conflating multiple subthreads, or too fine-grained, prematurely committing to a particular subcontext. This pattern also helps explain the granularity trade-off observed in \refsec{sec:ablation}. Flat labels are useful for filtering, but committing too early at the wrong level of abstraction can suppress relevant evidence before ranking or reasoning occurs. These findings suggest that future extensions should support hierarchical label spaces or defer label refinement until query time, rather than relying on irrevocable one-shot label selection. Appendix~\ref{app:error_analysis} provides the diagnostic setup, overlap-based statistics, and extended failure analysis.

\begin{table}[t]
\centering
\small
\setlength{\tabcolsep}{6pt}
\begin{tabular}{lc}
\toprule
\textbf{Failure Mode} & \textbf{Frequency (\%)} \\
\midrule
Non\_Inducible\_Label & 78.4 \\
Granularity\_Mismatch & 71.8 \\
\bottomrule
\end{tabular}
\caption{Distribution of failure modes in question-time label selection.
Percentages are computed over all judged $\langle$question, label type$\rangle$ cases.
A single case may exhibit multiple failure modes.}
\label{tab:label_failure_modes}
\end{table}

\subsection{Robustness to Configuration Choices}
\label{sec:robustness}
\paragraph{Sensitivity to Segmentation Granularity.}
We fix the thematic-scope window (\refsec{sec:thematic_scope}) to a domain-agnostic default of 50 turns, rather than tuning it for each benchmark setting. To characterize the effect of segmentation granularity, we evaluate one representative trajectory from each \benchmarkname{} partition under two stress settings: over-segmentation ($10$ turns) and under-segmentation ($100$ turns). Table~\ref{tab:granularity_sensitivity} reports Macro-F1 broken down by question type (\refsec{sec:benchmark_question_types}). The comparison reveals complementary failure modes at the two extremes. Under-segmentation most strongly harms Type~2 (Context-Aware Factual Recall), since distinct episodes are merged and near-duplicate facts become harder to disambiguate. Over-segmentation more noticeably reduces Type~3 (Context-Aware Multi-Hop Reasoning), because shorter windows fragment trajectory continuity and coreference chains. These results indicate that thematic-scope induction requires a balanced granularity. Windows that are too large blur episode boundaries, whereas windows that are too small disrupt the continuity needed for reasoning across turns.

\paragraph{Sensitivity to Backbone Choice.}
We further study whether the gains of \methodname{} depend on the choice of backbone model. Table~\ref{tab:backbone_sensitivity} compares \methodname{} with a full-context baseline and SeCom, the strongest structure-augmented RAG baseline, with each system instantiated using the same backbone. With gpt-4o-mini, \methodname{} outperforms both references on all three partitions, and the gap widens as context length increases. With gpt-4.1-mini, the full-context baseline remains strong on Small, but \methodname{} scales better, surpassing it on Medium and Large while consistently outperforming SeCom across all partitions. Overall, the improvement persists across backbone choices, indicating that the gains primarily come from the contextual-intent formulation rather than from a particular frontier model; stronger backbones further amplify the benefit.

%% file: tables/ablation_table.tex
\begin{table}[t]
\centering
\setlength{\tabcolsep}{5pt}
\resizebox{\columnwidth}{!}{%
\begin{tabular}{lcccccc}
\toprule
\multirow{2}{*}{\textbf{Variant}} & \multicolumn{6}{c}{\textbf{\benchmarkname}} \\
\cmidrule(lr){2-7}
 & $\text{Prec}_{S}$ & $\text{F1}_{S}$ & $\text{Prec}_{M}$ & $\text{F1}_{M}$ & $\text{Prec}_{L}$ & $\text{F1}_{L}$ \\
\midrule
\methodname{} (full) & \textbf{0.810} & \textbf{0.844} & \textbf{0.665} & \textbf{0.682} & \textbf{0.616} & \textbf{0.592} \\
\quad w/o thematic scope & 0.457 & 0.463 & 0.257 & 0.257 & 0.269 & 0.213 \\
\quad w/o event type & 0.730 & 0.753 & 0.525 & 0.527 & 0.383 & 0.273 \\
\quad w/o coreference & 0.554 & 0.578 & 0.508 & 0.489 & 0.469 & 0.404 \\
\quad w/o key entity type & 0.713 & 0.735 & 0.513 & 0.511 & 0.513 & 0.458 \\
\bottomrule
\end{tabular}%
}
\caption{Ablation results for \methodname{} on \benchmarkname{}. We report answer-set Macro Precision/Macro-F1 on \benchmarkname$_S$/$\!_M$/$\!_L$ ($N{=}144/168/61$).}
\label{tab:ablation_results}
\end{table}

%% file: sections/6_related_works.tex
\begin{figure}[t!]
    \centering
    % Using width=\columnwidth ensures the image fits within a single column.
    \includegraphics[width=\columnwidth]{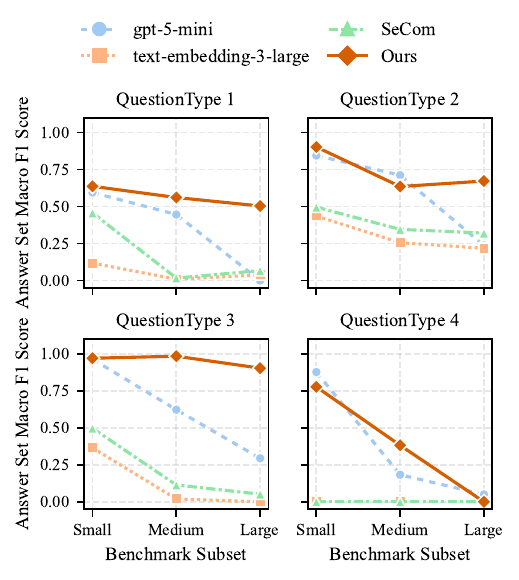}
    \caption{Results broken down by question type in \benchmarkname{}. We compare \methodname{} with the strongest baseline in each category: gpt-5-mini for Long-Context Models, text-embedding-3-large for Embedding RAG Agents, and SeCom~\citep{secom} for Structure-Augmented RAG Agents. The evaluation addresses four distinct capabilities: Incremental Memory Update (Type 1), Context-Aware Factual Recall (Type 2), Context-Aware Multi-Hop Reasoning (Type 3), and Information Synthesis (Type 4). See ~\refsec{sec:benchmark_question_types} for detailed definitions.}
    \label{fig:ablation_study_by_type}
\end{figure}

\paragraph{Memory-Augmented Agent Systems}
Existing memory architectures address the finite context window of LLMs by externalizing history via vector retrieval \cite{karpukhin-etal-2020-dense}, hierarchical summarization \cite{raptor, secom}, or graph-based storage \cite{graphrag, amem, hipporag}. While these methods expand recall capacity, they predominantly optimize for storage efficiency and semantic similarity rather than \emph{contextual validity}. Retrieval mechanisms driven solely by embedding alignment remain susceptible to noise when relevance depends on latent goals or evolving constraints. Furthermore, abstraction-based approaches often sacrifice essential details, while static schemas lack flexibility to adapt to shifting task demands. Consequently, current systems often fail to ground retrieved information in the agent's active intent, leaving a gap between access and effective state-dependent reasoning.

\paragraph{Long-Context Agentic Benchmarks}
Recent benchmarks such as LongMemEval \cite{longmemeval}, LongBench \cite{bai2024longbench2}, and LoCoMo \cite{locomo} have moved beyond single-turn QA to probe memory retention over extended horizons. While instrumental in revealing performance degradation in long-context settings, these evaluations frequently rely on structural oversimplifications that diverge from realistic workflows. Many employ strict turn-taking dynamics \cite{mtbench} or partition interactions into disjoint topics, allowing models to exploit local adjacency heuristics rather than performing robust context tracking. Unlike open-ended social dialogue \cite{locomo}, goal-oriented agent tasks require rigorous state maintenance across interleaved, non-local dependencies---a challenge that current benchmarks, with their focus on separable sub-tasks, fail to adequately capture.

%% file: sections/7_conclusion.tex
We present \textbf{\methodname{}}, a lightweight, domain-agnostic agentic memory system that models context history via \emph{contextual intent} and uses it to filter and prioritize memory for retrieval. This intent-aware cue suppresses semantically similar but contextually mismatched history, supporting robust long-horizon reasoning. We also introduce \textbf{\benchmarkname{}}, a benchmark that exposes the fragility of existing methods in realistic, goal-oriented trajectories. Across settings, \methodname{} achieves state-of-the-art performance and scales reliably as context grows, avoiding the degradation observed in long-context baselines on extended trajectories.

%% file: sections/8_limitation.tex
\methodname{} is designed to prioritize retrieval robustness and structural coherence in long-horizon contexts, utilizing a design philosophy that favors precision over raw ingestion speed. Consequently, the ingestion pipeline incurs multiple LLM calls to construct rich contextual intent tuples; while this overhead is amortized during retrieval to ensure superior disambiguation, the per-step ingestion cost is higher than lightweight embedding-only baselines. Similarly, our dynamic label evolution involves a trade-off between schema stability and immediate adaptability. While our system supports real-time updates, we employ a buffered update strategy (N=50) in our experiments to stabilize the label space. This choice ensures schema consistency but introduces a minor latency in formalizing novel event types that emerge in the middle of a buffer window. Finally, our induced event taxonomy is currently optimized for fine-grained factual lookup. While this granularity excels at specific queries, broad synthesis tasks may currently require aggregating across multiple event types. These trade-offs highlight promising avenues for future work, particularly in exploring hierarchical schema induction and lightweight structural predictors to further balance efficiency with our established performance gains.

%% file: sections/9_ethics.tex
We propose an entirely synthetic benchmark constructed within a closed-world environment. This design strictly prevents the inclusion of personally identifiable information and mitigates privacy risks associated with scraping real-world user data. In the Travel Planning domain, we utilize fictionalized entities to avoid potential harm to real-world businesses. For the Debate domain, we minimize the risk of generating toxic or harmful content by restricting topics to policy argumentation curated by the National Speech \& Debate Association and constraining agent generation to specific, retrieved evidence snippets.

Regarding human evaluation, we recruited volunteer annotators to validate trajectory quality and ground truth correctness. All participants provided informed consent, and no personal data was collected during the process. Finally, regarding the broader impact of agentic memory, our approach grounds memory in explicit contextual intent. This structural organization offers greater transparency and interpretability regarding retained information compared to black-box embedding methods, facilitating better auditability and user control over stored history.

%% file: sections/99_acknowledgement.tex
Research was supported in part by the AI Institute for Molecular Discovery, Synthetic Strategy, and Manufacturing: Molecule Maker Lab Institute (MMLI), funded by U.S. National Science Foundation under Award 2505932, NSF IIS 25-37827, and the Institute for Geospatial Understanding through an Integrative Discovery Environment (I-GUIDE) by NSF under Award No.\ 2118329. The research has used the Delta/DeltaAI advanced computing and data resource, supported in part by the University of Illinois Urbana-Champaign and through allocation \#250851 from the Advanced Cyberinfrastructure Coordination Ecosystem: Services \& Support (ACCESS) program, which is supported by National Science Foundation grants OAC 2320345, \#2138259, \#2138286, \#2138307, \#2137603, and \#2138296. We also thank Ling Wu and Haoyu Zhai for their helpful feedback.

%% file: sections/999_appendix.tex
\appendix
\etocdepthtag.toc{mtappendix}

\section*{Appendix Table of Contents}
\begingroup
  \etocsettagdepth{mtchapter}{none}
  \etocsettagdepth{mtappendix}{subsection}
  \etocsettocstyle{}{}
  \tableofcontents
\endgroup

\newpage

\input{sections/appendix/benchmark}

\input{sections/appendix/coreference_resolution}

\input{sections/appendix/baseline_description}

\input{sections/appendix/error_analysis}
\input{sections/appendix/cost}
\input{sections/appendix/robustness}
\input{sections/appendix/recall}
\input{sections/appendix/traj_number}

\section{LLM Prompt Specifications}
\input{sections/appendix/method_prompt}
\input{sections/appendix/evaluation_prompt}

\input{sections/appendix/trajectory_example}

\onecolumn

\input{examples/benchmark_trajectory_travel_example}
\input{examples/benchmark_trajectory_debate_example}
\begin{figure*}
\input{prompts/context_scope_generation}
\captionof{lstlisting}{Prompts for thematic scope generation, as detailed in \refsec{sec:thematic_scope}. In the prompt text, we refer to thematic scope as ``context scope,'' and ``prior structure notes'' correspond to the recent history $H_{\text{scope}}$.}
\label{lst:context_scope_generation}
\end{figure*}

\begin{figure*}
\input{prompts/canonical_summary_generation}
\captionof{lstlisting}{Prompts for generating a compressed summary $\Sigma_{\sigma}$  of thematic scope, as detailed in \refsec{sec:thematic_scope}.}
\label{lst:scope_summary}
\end{figure*}

\begin{figure*}
\input{prompts/event_type_generation}
\captionof{lstlisting}{Prompts for generating initial event label seeds and evolving the dynamic label space, as detailed in \refsec{sec:event_label}.}
\label{lst:event_labeling}
\end{figure*}

\begin{figure*}
\input{prompts/key_entity_type_generation}
\captionof{lstlisting}{Prompts for generating key entity type label seeds and evolving the dynamic label space, as detailed in \refsec{sec:entity_label}. In the prompt text, we refer to key entity types as ``function types.''}
\label{lst:entity_labeling}
\end{figure*}

\begin{figure*}
\input{prompts/memory_snippet}
\captionof{lstlisting}{Prompts for generating a canonical summary $c_t$ for each step in the trajectory as the final step of memory snippet construction, as detailed in \refsec{sec:memory_snippet}.}
\label{lst:memory_snippet_generation}
\end{figure*}

\begin{figure*}
\input{prompts/context_scope_selection}
\captionof{lstlisting}{Prompts for selecting thematic scope $\mathcal{S}_q$ during retrieval as detailed in \refsec{sec:retrieval}. In the prompt text, we refer to thematic scope as ``context scope''.}
\label{lst:thematic_scope_selection}
\end{figure*}

\begin{figure*}
\input{prompts/event_type_selection}
\captionof{lstlisting}{Prompts for selecting event type $\mathcal{E}_q$ during retrieval as detailed in \refsec{sec:retrieval}.}
\label{lst:event_label_selection}
\end{figure*}

\begin{figure*}
\input{prompts/key_entity_type_selection}
\captionof{lstlisting}{Prompts for selecting key entity types $\mathcal{K}_q$ during retrieval as detailed in \refsec{sec:retrieval}. In the prompt text, we refer to key entity types as ``function types.''}
\label{lst:key_entity_type_selection}
\end{figure*}

\begin{figure*}
\input{prompts/response_generation_general_travel}
\captionof{lstlisting}{Response generation prompts for open-domain and travel planning question answering in \benchmarkname.}
\label{lst:response_generation_travel}
\end{figure*}

\begin{figure*}
\input{prompts/response_generation_debate}
\captionof{lstlisting}{Response generation prompts for debate question answering in \benchmarkname.}
\label{lst:response_generation_debate}
\end{figure*}

\begin{figure*}
\input{prompts/answer_evaluator_single}
\captionof{lstlisting}{Answer automatic evaluation prompt with LLM for question answer that is a single answer}
\label{lst:answer_eval_single}
\end{figure*}

\begin{figure*}
\input{prompts/answer_evaluator_set}
\captionof{lstlisting}{Answer automatic evaluation prompt with LLM for question answer that is an answer set}
\label{lst:answer_eval_set}
\end{figure*}

%% file: sections/appendix/benchmark.tex
\section{Benchmark Construction and Implementation Details}
\label{sec:appendix_construction}

In this section, we provide comprehensive details regarding the \benchmarkname\ construction, implementation, and quality assurance protocols. We first detail the four-stage curation pipeline used to generate valid trajectories (\S\ref{app:curation_pipeline}). We then provide specific configurations for the \emph{Travel Planning} and \emph{Debate} domains (\S\ref{app:domain_specs}). Next, we describe the ground truth generation process and metric definitions (\S\ref{app:appendix_ground_truth}), followed by implementation statistics and reproducibility details (\S\ref{app:impl_stats}), and our quality assurance protocols (\S\ref{app:qa}). Finally, we include example trajectories and evaluation questions to illustrate the benchmark format (\S\ref{app:benchmark_example}).

\subsection{Curation Pipeline Details}
\label{app:curation_pipeline}

We employ a four-stage pipeline that decouples symbolic planning from natural-language generation.

\paragraph{Phase I: Closed-World Environment Construction.}
We first construct a closed-world environment $\mathcal{E}$ containing all static entities and knowledge required for each domain. This controlled environment ensures that agent actions remain grounded, avoids reliance on external tools, and mitigates data contamination. When trajectories require factual queries (e.g., hotel prices), we simulate and perturb real-world examples within $\mathcal{E}$ to avoid the data contamination issue \citep{sainz-etal-2023-nlp}; otherwise, we incorporate real-world information directly.

\paragraph{Phase II: Symbolic Storyboard Planning.}
In parallel, we construct symbolic storyboards capturing the high-level interaction plan. For each domain, we predefine an action space~$\mathcal{A}$ and specify pragmatic constraints on permissible action sequences. For \emph{Travel Planning}, we generate multi-day itinerary skeletons with slots for restaurants, accommodations, and attractions that require consistent state tracking. For \emph{Debate}, we employ a probabilistic argumentation planner that samples moves such as $\textsc{Attack}$, $\textsc{Defend}$, $\textsc{Concede}$, and $\textsc{Background}$ subject to dialogue rules (e.g., concessions require preceding successful attacks). Each action is paired with relevant information from the environment $\mathcal{E}$. To incorporate dense reuse of semantically similar content, the planner ensures some entities in $\mathcal{E}$ are referenced multiple times under different contextual intents.

This produces a symbolic trajectory $T_{\text{sym}} = \{\text{op}_1, \ldots, \text{op}_n\}$, where each operation $\text{op}_t$ specifies: (i) an agent role $r_t$, (ii) a latent goal $g_t$ describing potential actions and target entities, (iii) a payload $p_t$ containing domain-relevant details.

\paragraph{Phase III: Storyboard-Conditional Trajectory Generation.}
We translate symbolic storyboards into natural language using LLMs, producing surface-level actions that retain the structural constraints of the symbolic plan. For each operation $\text{op}_t$, the generator conditions on the role $r_t$, latent goal $g_t$, and payload $p_t$ to generate an action realization $a_t$ in natural language. We then assign a synthetic timestamp $\tau_t$ (e.g., a monotonically increasing time index) to obtain the trajectory step
$s_t = (r_t, a_t, \tau_t)$, yielding the natural-language trajectory $T = \{s_1, \ldots, s_n\}$.
A separate verification model checks entailment between the generated action $a_t$ and the symbolic specification of $\text{op}_t$ to ensure that the realization faithfully instantiates $g_t$ and $p_t$ under role $r_t$.

\paragraph{Phase IV: Pragmatic Refinement.}
We apply post-processing refinements that mimic natural communication patterns. First, we perform \textit{referential remodeling}, rewriting explicit entity mentions into complex referring expressions (e.g., ``Apollo Hotel'' becomes ``the second hotel I mentioned for Day~1''), forcing models to resolve coreference chains over long horizons. Second, we apply \textit{turn segmentation}, splitting monolithic generations into smaller discourse units to simulate realistic conversational dynamics. Third, for entities in $\mathcal{E}$ that can be easily disambiguated via surface semantics, we perform \textit{entity normalization}, masking semantic content (e.g., replacing explicit evidence sources with author--year citations in debate settings).

\subsection{Domain Specifications}
\label{app:domain_specs}

\subsubsection{Travel Planning Domain}
\label{app:travel_specs}
For the \emph{Travel Planning} domain, we construct a closed-world environment, $\mathcal{E}_{travel}$, consisting of 100 entities for each category: accommodations, restaurants, and attractions. Each entity possesses canonical descriptions and structured attributes (e.g., price tiers, ratings, cuisine types). This environment $\mathcal{E}_{travel}$ is shared across all travel planning trajectories.

To ensure distinctiveness, all entity names are adapted from ancient Greek mythology (e.g., \textit{Menelaus Spartan Grill}), while pricing and ratings are sampled from plausible real-world ranges.

For the symbolic trajectory $T_\text{sym}$, we define an action space $\mathcal{A}_{travel}$ which includes the following operations: \texttt{indicate\_date}, \texttt{propose\_option}, \texttt{inquire\_details}, \texttt{compare\_options}, and \texttt{make\_decision}. We enforce pragmatic constraints to ensure logical coherence: for instance, a \texttt{propose\_option} action must precede any \texttt{inquire} or \texttt{decision} actions, and a \texttt{compare\_options} action requires prior proposals to reference.

\subsubsection{Debate Domain}
\label{app:debate_specs}
For the \emph{Debate} domain,\footnote{We derive our debate topics from the National Speech \& Debate Association: \url{https://www.speechanddebate.org/}} a unique closed-world environment $\mathcal{E}_{debate}^{topic_i}$ is generated for each topic. This is achieved by retrieving knowledge snippets that serve as evidence supporting either side of the motion.\footnote{We utilize \url{serper.dev} as our internet retriever.} On average, the environment contains 50 unique, authoritative and realistic knowledge snippets for the small and medium partitions, and 200 for the large partition.

To enforce our \emph{entity normalization} mechanism, each evidence snippet used in a debate trajectory is cited inline as \textit{Author (Year)} to mask semantic meaning.

For the symbolic trajectory $T_\text{sym}$, we define an action space $\mathcal{A}_{debate}$ including: \texttt{propose\_argument}, \texttt{attack}, \texttt{defend}, \texttt{concede}, \texttt{supply\_background}, and \texttt{summarize}. The pragmatic constraints dictate that:
\begin{itemize}
    \item A \texttt{propose\_argument} operation must precede any \texttt{attack}, \texttt{defend}, or \texttt{concede} actions.
    \item A \texttt{concede} action must occur only after both an \texttt{attack} and a \texttt{defend} action have taken place.
    \item The \texttt{summarize} action must be the terminal operation in $T_\text{sym}$.
\end{itemize}

\subsection{Ground Truth Generation and Metrics}
\label{app:appendix_ground_truth}

\paragraph{Ground Truth Answer Generation.}
Ground-truth answers are derived deterministically from the symbolic trajectory operations $T_{\text{sym}}$ defined in Phase II. For each question $q \in Q$, we identify the subset of operations $Q_{\text{ops}}(q) = \{\text{op}_{i_1}, \ldots, \text{op}_{i_k}\}$ whose payloads contain the requisite information. These symbolic operations are then processed by an LLM (\texttt{gpt-4o-mini}) to generate natural-language answers. Depending on the question type, the ground truth may be a single information snippet, a set of entities, or a structured response. To ensure correctness, we sample a subset of generated answers and validate them through human annotation.

\paragraph{Evaluation Metric.}
Because ground-truth answers may consist of single items or sets of entities, we employ answer-set macro-averaged precision, recall, and F1 score as our primary evaluation metrics. For each question, we compute precision and recall by comparing the predicted answer set against the ground truth, derive the F1 score, and average across all questions. This metric provides a more nuanced assessment than binary accuracy, particularly for questions with multi-element answers, and better captures partial credit scenarios.

\subsection{Implementation Details}
\label{app:impl_stats}

\paragraph{Models used.}
Across benchmark construction phases (environment construction, storyboard-to-trajectory realization, and pragmatic refinement), we use gpt-5-mini with \texttt{temperature=1.0} and \texttt{reasoning\_effort=medium}. In the Travel Planning domain, surface utterances are generated with gpt-4.1-mini. In the Debate domain, evidence processing (query generation and evidence naming) uses gpt-4.1-mini, while debate utterances are generated with gpt-5-mini.

\paragraph{Prompts, label ontologies, and hyperparameters.}
We release all prompts used for action rendering, referential remodeling, turn segmentation, and any automated annotation used in the benchmark (see Appendix~\ref{app:stitch-prompts}). We also release the label schema used by \methodname{} (scope/event/target and optional functional-type), along with all pipeline hyperparameters (e.g., history window sizes, sampling budgets, retrieval caps, and truncation rules) required to reproduce the reported results.

\paragraph{Determinism.}
LLM outputs are not guaranteed deterministic; we enforce reproducibility where possible via (i) fixed random seeds in non-LLM sampling logic (e.g., itinerary sampling and storyboard planning), and (ii) stable processing order and deterministic identifiers (e.g., stable mapping for evidence IDs). We log all configuration files and random seeds used for each run.

\subsection{Benchmark Statistics.}
\label{app:benchmark_stats}

To investigate and evaluate LLM agent memory across varying context lengths, following \citet{longmemeval}, we construct three subsets of the benchmark, denoted as $\text{\benchmarkname}_{S}$, $\text{\benchmarkname}_{M}$, and $\text{\benchmarkname}_{L}$ representing small, medium, and large partitions as three scales, respectively. The statistics for these partitions are detailed in Table~\ref{tab:benchmark_stats}. Notably, due to the extreme context length ($>$400k tokens) and substantial computational cost of generating coherent long-horizon interactions, the Large subset comprises 2 high-density trajectories. Despite the lower trajectory count, this subset contains 61 evaluation questions, providing sufficient granularity to probe memory stability at scale.

\begin{table}[ht]
    \centering
    \caption{Statistics for the \benchmarkname{} partitions. \textit{Context Len.} denotes the average context length in tokens; \textit{Trajs.} indicates the number of unique trajectories; and \textit{Ques.} denotes the total number of evaluation questions.}
    \label{tab:benchmark_stats}
    \resizebox{\columnwidth}{!}{%
    \begin{tabular}{lccc}
        \toprule
        \textbf{Subset} & \textbf{Context Len.} & \textbf{Trajs.} & \textbf{Ques.} \\
        \midrule
        $\text{\benchmarkname}_{S}$ & 23k  & 6 & 144 \\
        $\text{\benchmarkname}_{M}$ & 137k & 6 & 168 \\
        $\text{\benchmarkname}_{L}$ & 408k & 2 & 61 \\
        \bottomrule
    \end{tabular}%
    }
\end{table}

\subsection{Quality Assurance and Verification}
\label{app:qa}

To ensure the quality and correctness of the benchmark, we employ a two-stage verification process involving both automatic checks and human annotation.

\paragraph{Automatic Verification.}
We utilize an LLM (gpt-4.1-mini) as a judge with \texttt{temperature=0.0} and \texttt{top\_p=0.9} to automatically verify the entailment between each symbolic operation $op_{t} \in T_{sym}$ and its natural language realization $\tilde{s}_t$. We randomly sampled 200 pairs of $(op_t, \tilde{s}_t)$ for evaluation, consisting of 100 samples from the \emph{Travel Planning} domain and 100 from the \emph{Debate} domain. The LLM-as-judge evaluation indicates a success rate of 99.5\%.

\paragraph{Human Verification.}
To assess the overall trajectory quality and the validity of the automatically generated questions and ground truth answers, we measured agreement with human annotators. Given the scale of the benchmark, we randomly sampled 2 $\text{\benchmarkname}_{s}$ trajectories from each domain (4 trajectories total). We recruited four volunteer human annotators with a master's degree or higher in computer science, or at least two years of experience in the AI/ML industry. Each trajectory was assigned to 2 annotators. Annotators were instructed to read the full trajectory and answer the free-response evaluation questions. The authors then manually verified the correctness of the human answers against the LLM-generated ground truth. On average, human annotators agreed with 98.25\% of the LLM-generated ground truth answers.

\paragraph{Instructions Provided to Annotators.}
\begin{quote}
Please read the entire trajectory carefully before answering any questions.

\begin{itemize}[leftmargin=*]
    \item Each question should be answered using only the information explicitly stated or logically implied in the trajectory.
    \item Do \emph{not} rely on external knowledge, background assumptions, or real-world facts that are not supported by the trajectory.
    \item Provide a concise, free-form textual answer for each question.
    \item If the answer is not present in the trajectory, is ambiguous, or cannot be determined with certainty, respond exactly with: \texttt{cannot be answered}.
    \item Do not guess or infer missing information. When multiple interpretations are possible, choose \texttt{cannot be answered}.
\end{itemize}
\end{quote}

All annotators received identical instructions and completed the task independently.

\subsection{Example Trajectories and Evaluation Questions}
\label{app:benchmark_example}
We provide partial trajectory examples from the Travel Planning and Debate domains in Appendix~\ref{app:travel_trajectory} and Appendix~\ref{app:debate_trajectory} respectively. Additionally, we present examples for each of the four evaluation question types below.

\input{examples/benchmark_question_example}

%% file: examples/benchmark_question_example.tex
% --- Type 1 ---
\questionblock{type1bg}{Type 1: State Tracking}{%
"What is the final hotel chosen for Day 1? Provide the hotel's name." \\
"List all restaurant(s) that were decided or chosen at any point for Day 1’s lunch, including intermediate and final decisions."
}

% --- Type 2 ---
\questionblock{type2bg}{Type 2: Context-Aware Factual Recall}{%
"List the evidence used by the con side during the debate. Provide the evidence names separated by semicolons."
}

% --- Type 3 ---
\questionblock{type3bg}{Type 3: Context-Aware Multi-Hop Reasoning}{%
"Which attractions were discussed for Day 1's attraction? List all names."
}

% --- Type 4 ---
\questionblock{type4bg}{Type 4: Information Synthesis}{%
"What is the final travel plan for Day 1? Answer in the following format: Stay at [accommodation to be filled]; Breakfast at [breakfast to be filled]; Lunch at [lunch to be filled]; Dinner at [dinner to be filled]; Visit [attraction(s) to be filled]"
}

%% file: sections/appendix/coreference_resolution.tex
\section{Verification of Structural Alignment and Coreference Resolution}
\label{app:coref_verification}

In Section~\ref{subsec:construction}, we describe a module for \textbf{Coreference Resolution via Structural Alignment}, where ambiguous steps (e.g., \textit{``Book it''}) are rewritten into explicit forms (e.g., \textit{``Book the Sheraton''}) using the inferred contextual intent. To rigorously quantify the success of this module, we developed a deterministic verification pipeline that measures \textbf{Entity Resolution Recall (ERR)}.

\subsection{Metric Definition}
We define Entity Resolution Recall as the proportion of ground-truth entities (implicit in the user's intent or explicitly tracked in the simulation state) that are successfully surfaced in the final textual representation of the memory snippet.

Let $T_{\text{gold}}$ be the set of symbolic ground-truth operations in the trajectory, where each operation $op_i$ is associated with a specific target entity $e_i$ (e.g., a specific hotel name or evidence document). Let $M_{\text{gen}}$ be the set of generated memory snippets, where $m_i \in M_{\text{gen}}$ denotes the snippet corresponding to the timestamp of operation $op_i$. We define the validation function $V(e_i, m_i) \to \{0, 1\}$ as a strict string-matching operation. The ERR is calculated as:

\begin{equation}
\text{ERR} = \frac{\sum_{i=1}^{|T_{\text{gold}}|} V(e_i, m_i)}{|T_{\text{gold}}|}
\end{equation}

\subsection{Implementation Details}
To avoid human bias in evaluation, we implemented an automated verification suite tailored to each domain's specific data structure.

\paragraph{Travel Planning Domain.}
In the Travel Planning dataset, the ground-truth simulation log contains structured payloads for actions such as \texttt{PROPOSE}, \texttt{DECIDE}, and \texttt{COMPARE}. Each payload includes a canonical \texttt{entity} (e.g., \textit{``Apollo Hotel''}). We align generated notes to simulation turns via discrete time steps ($t$). Both the ground-truth \texttt{entity} and the generated note text undergo normalization (lowercasing, whitespace trimming). For every turn involving a referential action, the script asserts that the canonical \texttt{entity} appears as a substring in the rewritten memory snippet.

\paragraph{Debate Domain.}
In the Debate dataset, coreference often involves referring to evidence documents (e.g., \textit{``that study we discussed''}). We maintain a closed-world registry mapping every evidence UUID to its \texttt{neutral\_name} (e.g., \textit{``Smith et al., 2023 Study on Microplastics''}). For any turn where the agent utilizes evidence (indicated by \texttt{action\_object} or \texttt{evidence\_used} metadata), we verify that the corresponding \texttt{neutral\_name} is explicitly present in the memory snippet.

\subsection{Results}
We conducted this audit across three scales of trajectory complexity (Small, Medium, Large) for both domains. Table~\ref{tab:coref_results} summarizes the Entity Resolution Recall. The system demonstrates high resolution in short contexts and maintains robust performance (approximately $80\%$) even as trajectory length and complexity increase.

\begin{table}[h]
    \centering
    \caption{Entity Resolution Recall (ERR) across domains and trajectory scales. Scores represent the average percentage of implicit references successfully rewritten into explicit entity mentions.}
    \label{tab:coref_results}
    \begin{tabular}{llc}
        \toprule
        \textbf{Domain} & \textbf{Scale} & \textbf{Recall (Avg)} \\
        \midrule
        \multirow{3}{*}{Travel Planning} & Small & 97.4\% \\
                                         & Medium & 79.2\% \\
                                         & Large & 81.1\% \\
        \midrule
        \multirow{3}{*}{Debate} & Small & 91.5\% \\
                                & Medium & 78.6\% \\
                                & Large & 81.1\% \\
        \bottomrule
    \end{tabular}
\end{table}

%% file: sections/appendix/baseline_description.tex
\section{Baseline Descriptions}
\label{app:baseline_description}

\textbf{RAPTOR} \citep{raptor} proposes a tree-structured retrieval system that recursively summarizes text chunks to capture information at varying levels of abstraction. It constructs a hierarchical tree where leaf nodes represent raw text segments and higher-level nodes represent summarized clusters of child nodes. During inference, RAPTOR retrieves from this multi-layered tree, allowing retrieval at different levels of abstraction, supporting both high-level thematic queries and fine-grained factual queries.

\textbf{GraphRAG} \citep{graphrag} is a structured retrieval approach that utilizes Large Language Models (LLMs) to extract a knowledge graph from the corpus, identifying entities and their relationships. Unlike traditional graph methods that rely on strict schema matching, GraphRAG employs community detection algorithms (e.g., Leiden) to partition the graph into modular communities and generates natural language summaries for each community. Retrieval involves matching the query against community-level summaries to support global reasoning, often complemented by local entity-level retrieval.

\textbf{HippoRAG 2} \citep{hipporag} is a neurobiologically inspired retrieval framework that mimics the hippocampal indexing theory of human memory. It constructs a knowledge graph where nodes serve as associative indices rather than direct factual stores. During retrieval, HippoRAG identifies key entities in the query and uses the Personalized PageRank (PPR) algorithm to spread activation across the graph, identifying paths to relevant information even when there is no direct lexical or semantic overlap. This allows the system to perform multi-hop reasoning by surfacing contextually related passages that are separated in the raw text but connected via shared or related entities in the induced knowledge graph.

\textbf{A-mem} \citep{amem} introduces an agentic memory system inspired by the Zettelkasten note-taking method. It organizes interaction history into atomic, interconnected "notes" rather than a flat list of turns. A key innovation is its \emph{memory evolution} mechanism: when new information is acquired, the system not only generates new notes but also dynamically updates and refines existing notes and their links. This allows the memory to self-organize and adapt over long horizons, supporting retrieval that combines semantic matching with graph-based contextual expansion over interconnected notes.

\textbf{SeCom} \citep{secom} focuses on optimizing the granularity of memory units for long-term conversations. The authors argue that traditional turn-level (too fine) and session-level (too coarse) retrieval are suboptimal. Instead, SeCom introduces a \emph{segment-level} approach, using a specialized model to partition conversation history into discourse-coherent segments. Furthermore, it applies compression-based denoising (via LLMLingua-2) to these segments to remove redundancy. This ensures that the retrieval mechanism targets semantically complete discourse units without being distracted by the noise inherent in raw conversational transcripts.

%% file: sections/appendix/error_analysis.tex
\section{Error Analysis}
\label{app:error_analysis}
This section expands the concise error analysis in the main text with full diagnostics, methodology details, and additional qualitative discussion.
\subsection{Question-Side Label Selection}
\label{app:error_analysis_label_overlap}

To better understand the role and limitations of the question-side label selection component in our retrieval pipeline, we conduct an error analysis that isolates this step from downstream retrieval and generation. The goal of this analysis is not to evaluate end-task performance, but to characterize how well the labels selected for a question align with the labels assigned to ground-truth (GT) turns in the dialogue history.

\paragraph{Methodology.}
For each question, our method selects a set of labels spanning multiple types (e.g., context scope, event type, target, and key entity). Each candidate turn in the trajectory, including all GT turns, is annotated with structured labels of the same types. We then compute the overlap between the question’s selected labels and each candidate turn’s labels. This overlap-based analysis evaluates the intrinsic discriminative signal provided by the question-side label selection, independent of any retrieval model.

We derive several normalized statistics at the question and turn levels. At the question level, we measure whether at least one GT turn attains the maximum label overlap among all candidate turns (\texttt{Q\_GT\_AT\_MAX}), whether any GT turn fully covers the selected labels (\texttt{Q\_GT\_FULL}), and whether all candidate turns have zero overlap (\texttt{Q\_ZERO}). At the turn level, we compute the fraction of GT turns that achieve the maximum overlap (\texttt{GT\_DOM}) as well as the average fraction of selected labels supported by GT turns (\texttt{GT\_COV}). We additionally report optional diagnostics capturing cases where non-GT turns strictly dominate GT turns in overlap, and the best achievable GT coverage per question. All statistics are first computed per dataset instance and then macro-averaged across datasets within each domain. Table~\ref{tab:question_side_label_overlap} shows detailed statistics.

\paragraph{Overall Findings.}
Across both debate and travel planning datasets, question-side label selection consistently provides non-zero overlap with candidate turns, as reflected by a near-zero \texttt{Q\_ZERO} rate. This indicates that the selected labels are generally relevant to the dialogue trajectory and introduce meaningful constraints.

At the same time, the analysis reveals clear limits in sufficiency. No dataset exhibits full coverage of the selected labels by any single GT turn (\texttt{Q\_GT\_FULL} = 0 across all settings), and average GT coverage remains well below full alignment. Even when GT turns reach the maximum overlap, they typically support only a subset of the selected labels. This pattern suggests that the question-side label selection often encodes constraints that are not jointly realized within a single turn.

Furthermore, overlap density alone is not a reliable indicator of correctness. The \texttt{GT\_DOM} metric shows that GT turns rarely dominate the candidate set by overlap, and in a substantial fraction of questions, non-GT turns achieve strictly higher overlap than all GT turns. This implies that while selected labels help surface relevant regions of the trajectory, they do not, by themselves, uniquely identify the correct turn.

\paragraph{Label-Type Behavior.}
Breaking down the analysis by label type provides additional insight. Context-scope labels frequently enable GT turns to reach the maximum overlap, indicating strong recall-oriented behavior, but their coverage remains low, reflecting coarse granularity. Key entity labels achieve higher coverage when they align with the question, but exhibit greater variance, indicating sensitivity to whether the question provides sufficient cues for reliable selection. Event and target labels offer intermediate behavior, contributing useful signal but rarely achieving high coverage in isolation. These results indicate that no single label type is sufficient on its own, and that their contributions are complementary but partial.

\paragraph{Implications.}
Taken together, these findings show that question-side label selection plays an important role in structuring the retrieval space, but is not sufficient as a standalone mechanism for precise turn identification. The selected labels capture aspects of the question’s intent and reduce the search space, yet they frequently introduce constraints that are either overly coarse or overly specific relative to the information localized in individual turns. This motivates the need for downstream mechanisms that can reason over partial label alignment, rather than relying solely on flat overlap counts.

Importantly, they clarify the scope of what question-side label selection can reliably provide, and highlight opportunities for improving how label signals are combined and interpreted in long-horizon retrieval settings.

\begin{table}[t]
\centering
\small
\setlength{\tabcolsep}{4pt}
\begin{tabular}{lcc}
\toprule
\textbf{Metric} & \textbf{Debate} & \textbf{Travel} \\
\midrule
Q\_GT@MAX        & $0.50 \pm 0.14$ & $0.66 \pm 0.08$ \\
GT\_DOM          & $0.14 \pm 0.07$ & $0.29 \pm 0.08$ \\
GT\_COV          & $0.15 \pm 0.05$ & $0.25 \pm 0.03$ \\
GT\_BEST\_COV    & $0.22 \pm 0.06$ & $0.33 \pm 0.06$ \\
Q\_ZERO          & $0.00$          & $0.00$          \\
\bottomrule
\end{tabular}
\caption{Question-side label selection diagnostics (mean $\pm$ std across datasets).
Q\_GT@MAX: fraction of questions where at least one GT turn attains maximum overlap.
GT\_DOM: fraction of GT turns that are densest matches.
GT\_COV / GT\_BEST\_COV: average and best-case coverage of selected labels by GT turns.
Q\_ZERO: fraction of questions with zero overlap across all candidates.}
\label{tab:question_side_label_overlap}
\end{table}

\subsection{Failure Modes in Question-Time Label Selection}
\label{sec:error_analysis_label_failure}

In addition to overlap-based diagnostics, we conduct a targeted error analysis to characterize \emph{why} question-time label selection strongly determines the correctness of retrieval. This analysis focuses exclusively on the label selection step and does not depend on retrieved content or generation quality.

\paragraph{Methodology.}
We analyze questions that lead to incorrect answers under otherwise successful execution, and evaluate the labels selected at question time for each label type (context scope, event type, target, and functional seed). For each such \textlangle question, label type \textrangle{} pair, we invoke an LLM-based judge to assess whether the selected labels are appropriate given the question alone. Importantly, the judge is explicitly instructed to rely only on information that can be justified from the question, without assuming access to the dialogue trajectory or downstream retrieval results.

The judge diagnoses failures using a fixed taxonomy with three categories: \textbf{Non\_Inducible\_Label} and \textbf{Granularity\_Mismatch}. Each judgment is accompanied by a rationale and a confidence score. In this section, we focus on the two most prevalent and structurally consequential error types.

\paragraph{Non-Inducible Label Selection.}
The most frequent failure mode is Non\_Inducible\_Label, accounting for 78.4\% of all observed label selection errors across label types. In these cases, the model selects labels that cannot be justified from the question by explicit lexical cues, clear paraphrase, or necessary semantic entailment. Despite prompt instructions encouraging conservative selection, the model frequently introduces labels that reflect plausible but ungrounded abstractions.

This behavior reflects a fundamental asymmetry in the label selection setting. Candidate labels are derived from full dialogue trajectories and therefore encode rich structural regularities about what commonly occurs in such interactions. In contrast, the question specifies only a narrow and partial information need. When exposed to trajectory-derived label inventories, the model tends to project these latent regularities onto the question, selecting labels that are coherent in the broader conversational context but not inducible from the question itself. As a result, label selection is biased toward what is \emph{typical} of the trajectory rather than what is \emph{licensed} by the question.

These findings indicate that prompt-level constraints alone are insufficient to reliably prevent non-inducible label selection once the model is exposed to trajectory-derived abstractions.

\paragraph{Granularity Mismatch.}
The second dominant failure mode is Granularity\_Mismatch, which occurs in 71.8\% of label selection errors. This error arises when the abstraction level of selected labels does not align with the scope implied by the question. Overly coarse labels conflate multiple subthreads or contexts, while overly fine-grained labels prematurely commit to specific subcontexts that the question does not warrant.

This failure mode has direct implications for retrieval. Because turn retrieval is strictly gated by the selected labels, a granularity mismatch can exclude relevant turns or admit large numbers of irrelevant ones before any content is examined. Crucially, this error occurs \emph{prior to retrieval}, leaving no opportunity for downstream correction through ranking or reasoning. From the perspective of question-time selection, the model is forced to commit to a single level of abstraction despite uncertainty about the question’s true scope.

The prevalence of granularity mismatch highlights a limitation of flat label spaces: without hierarchical organization, label selection must resolve abstraction level prematurely, increasing the risk of irreversible filtering errors.

\paragraph{Implications.}
Together, these findings clarify the role of question-time label selection in long-horizon retrieval, exposing intrinsic challenges in selecting trajectory-derived abstractions from underspecified questions. They motivate extensions that allow label signals to be refined, deferred, or adjusted dynamically, rather than committing irrevocably at question time.

\subsection{Error Analysis on LoCoMo}
To better understand why STITCH is strongest on CAME-Bench and LongMemEval yet only competitive on LoCoMo, we manually analyzed 60 randomly sampled LoCoMo failure cases. The dominant failure mode was contextual-intent coverage mismatch (53.3\%). Many LoCoMo questions probe open-ended conversational details that are only weakly coupled to explicit goals or task states, whereas STITCH organizes memory around thematic scope, event type, and key entity types that are optimized for goal- and task-state tracking. When the queried information is socially salient but only loosely connected to a persistent goal trajectory, the induced structure may not provide sufficiently discriminative retrieval cues.
We identified three secondary failure modes. First, label-density ranking errors (16.7\%) arose when relevant snippets were retrieved but deprioritized because the correct evidence had lower overlap density than distractors. Second, unresolved information conflicts (21.7\%) occurred when the dialogue contained conflicting or updated facts and the abstraction step failed to consistently prioritize the latest or most authoritative information. Third, recontextualization/linkage loss (8.3\%) occurred when an answer depended on subtle conversational framing that was weakened during snippet rewriting or summarization. These findings reinforce the intended scope of STITCH: the method is most effective when memory retrieval must preserve contextual validity under interleaved, long-horizon goal evolution, while LoCoMo probes a partially overlapping but distinct conversational-memory regime.

%% file: sections/appendix/cost.tex
\section{Token Cost Analysis}

To complement the main performance results, we report a model-agnostic token cost breakdown and separate \emph{ingestion} from \emph{query} cost. This distinction is important for agent memory systems because memory must be constructed \emph{online}: the system observes a streaming trajectory, incrementally updates its external memory, and must support retrieval at arbitrary intermediate times. As a result, an agentic memory method cannot assume access to the full trajectory only at the end and then retroactively optimize how history is stored. This accounting matches the design of \methodname{}, which first performs online contextual-intent construction during ingestion and later performs intent-aware retrieval at question time.

For fair comparison, we adapt SeCom to the same streaming setting and denote this variant as SeCom$^\ast$. Under our accounting, ingestion tokens measure the cost of transforming a streaming trajectory into an external memory representation, while query tokens measure the cost of question-time retrieval and answer generation. Table~\ref{tab:cost_breakdown} reports the resulting token counts on CAME-Bench S/M/L. Although \methodname{} incurs additional ingestion cost relative to lightweight baselines, it consistently requires fewer query-time tokens than strong structured-memory baselines. This behavior reflects the intended trade-off discussed in our limitations: \methodname{} invests more computation upfront to construct richer memory snippets, and later amortizes that cost through cheaper, more targeted retrieval.

\begin{table}[t]
\centering
\scriptsize
\setlength{\tabcolsep}{3.5pt}
\begin{tabular}{@{}lcccccc@{}}
\toprule
& \multicolumn{3}{c}{Ingestion Tok. (K)} & \multicolumn{3}{c}{Query Tok. (K)} \\
\cmidrule(lr){2-4} \cmidrule(lr){5-7}
Method & S & M & L & S & M & L \\
\midrule
GPT-5-mini & -- & -- & -- & 26.8 & 199.9 & 266.1 \\
\shortstack[l]{text-embedding-\\3-large} & -- & -- & -- & 3.5 & 3.2 & 3.1 \\
SeCom$^\ast$ & 5.2 & 5.5 & 6.0 & 6.0 & 11.1 & 6.1 \\
\methodname{} & 6.1 & 3.9 & 3.2 & \textbf{1.4} & \textbf{3.2} & \textbf{1.9} \\
\bottomrule
\end{tabular}
\caption{Model-agnostic token cost breakdown on CAME-Bench. Values are reported in thousands of tokens. Ingestion tokens measure online memory construction from a streaming trajectory; query tokens measure question-time retrieval and answer generation. SeCom$^\ast$ is our streaming adaptation of SeCom; ``--'' indicates no separate online memory-construction phase under this accounting.}
\label{tab:cost_breakdown}
\end{table}

%% file: sections/appendix/robustness.tex
\section{Additional Robustness Results}

We report two additional robustness analyses for \methodname{}: sensitivity to thematic-scope segmentation granularity and sensitivity to the choice of backbone model. These results complement the discussion in ~\refsec{sec:robustness} by providing the full tabulated comparisons.

\subsection{Segmentation Granularity}

To assess how sensitive \methodname{} is to the granularity of thematic-scope induction, we vary the scope window size around the default setting of 50 turns and evaluate two stress settings: over-segmentation (10 turns) and under-segmentation (100 turns). Table~\ref{tab:granularity_sensitivity} reports Macro-F1 broken down by question type for one representative trajectory from each CAME-Bench partition. Consistent with the main-text discussion, the results show a clear trade-off: overly large windows blur episode boundaries and most strongly hurt context-aware factual recall, while overly small windows fragment trajectory continuity and more strongly affect multi-hop reasoning.

\input{tables/granularity_sensitivity_table}

\subsection{Backbone Sensitivity}

To determine whether the gains of \methodname{} depend on a particular backbone, we compare \methodname{} against a full-context baseline and SeCom under matched backbone models. Table~\ref{tab:backbone_sensitivity} reports Macro Precision and Macro-F1 across CAME-Bench partitions. The improvements of \methodname{} persist across backbone choices, indicating that the gains primarily arise from the contextual-intent formulation rather than from a single frontier model, while stronger backbones further amplify the benefit.

\input{tables/backbone_sensitivity_table}

%% file: tables/granularity_sensitivity_table.tex
\begin{table}[t]
\centering
\setlength{\tabcolsep}{4pt}
\resizebox{\columnwidth}{!}{%
\begin{tabular}{llcccc}
\toprule
\textbf{Partition} & \textbf{Window} & \textbf{T1} & \textbf{T2} & \textbf{T3} & \textbf{T4} \\
\midrule
\multirow{3}{*}{Small} 
& 10 & 0.269 & 0.533 & 0.620 & 0.200 \\
& 50 (default) & 0.749 & 1.000 & 1.000 & 1.000 \\
& 100 & 0.222 & 0.267 & 0.646 & 0.200 \\
\midrule
\multirow{3}{*}{Medium} 
& 10 & 0.367 & 0.233 & 0.436 & 0.000 \\
& 50 (default) & 0.759 & 0.533 & 0.983 & 0.439 \\
& 100 & 0.298 & 0.000 & 0.626 & 0.025 \\
\midrule
\multirow{3}{*}{Large} 
& 10 & 0.120 & 0.467 & 0.558 & 0.017 \\
& 50 (default) & 0.504 & 0.800 & 0.904 & 0.000 \\
& 100 & 0.064 & 0.267 & 0.451 & 0.000 \\
\bottomrule
\end{tabular}%
}
\caption{Sensitivity of \methodname{} to the thematic-scope window size. We evaluate one representative trajectory from each \benchmarkname{} partition and report Macro-F1 by question type (\refsec{sec:benchmark_question_types}).}
\label{tab:granularity_sensitivity}
\end{table}

%% file: tables/backbone_sensitivity_table.tex
\begin{table}[t]
\centering
\setlength{\tabcolsep}{3pt}
\resizebox{\columnwidth}{!}{%
\begin{tabular}{lcccccc}
\toprule
\multirow{2}{*}{\textbf{Model}} & \multicolumn{2}{c}{\textbf{Small}} & \multicolumn{2}{c}{\textbf{Medium}} & \multicolumn{2}{c}{\textbf{Large}} \\
\cmidrule(lr){2-3} \cmidrule(lr){4-5} \cmidrule(lr){6-7}
 & \textbf{Prec.} & \textbf{F1} & \textbf{Prec.} & \textbf{F1} & \textbf{Prec.} & \textbf{F1} \\
\midrule
Full ctx. + GPT-4o-mini & 0.336 & 0.315 & 0.082 & 0.075 & 0.017 & 0.017 \\
SeCom + GPT-4o-mini & 0.463 & 0.458 & 0.227 & 0.219 & 0.238 & 0.222 \\
\methodname{} + GPT-4o-mini & 0.544 & 0.561 & 0.394 & 0.397 & 0.346 & 0.346 \\
Full ctx. + GPT-4.1-mini & 0.715 & 0.712 & 0.401 & 0.362 & 0.317 & 0.213 \\
SeCom + GPT-4.1-mini & 0.463 & 0.458 & 0.229 & 0.219 & 0.238 & 0.222 \\
\methodname{} + GPT-4.1-mini & 0.545 & 0.555 & 0.420 & 0.440 & 0.319 & 0.308 \\
\methodname{} + GPT-5-mini & 0.810 & 0.844 & 0.716 & 0.726 & 0.612 & 0.602 \\
\bottomrule
\end{tabular}%
}
\caption{Backbone sensitivity on \benchmarkname{}. We report Macro Precision and Macro-F1 across partitions for \methodname{}, a full-context baseline, and SeCom under matched backbone models. `Full ctx.' denotes supplying the full trajectory to the long-context model without structured memory.}
\label{tab:backbone_sensitivity}
\end{table}

%% file: sections/appendix/recall.tex
\section{Retrieval Recall Analysis}

We prioritize end-to-end evaluation in the main paper because gold retrieval is often underdetermined: many questions admit multiple valid evidence sets. For example, in the travel planning domain, a question such as ``What candidate accommodation options did we discuss?'' may be answerable from several distinct mentions of the same hotel, including an initial brainstorming turn, a later comparison of pros and cons, or a final decision turn. Any one of these mentions may be sufficient evidence, making a unique gold passage set inherently ambiguous.

For transparency, we nevertheless report retrieval recall where valid memories can be mapped back to raw trajectory turns. For methods whose memory representations cannot be reliably aligned to raw text, we omit them from this retrieval-level analysis. Table~\ref{tab:retrieval_recall} summarizes the resulting recall on CAME-Bench S/M/L. While several baselines attain higher recall on the Small partition, \methodname{} scales more favorably with context length and achieves the strongest recall on the Medium and Large partitions. This trend is consistent with the end-to-end results in Table~\ref{tab:retrieval_recall}: retrieval recall alone does not determine final answer quality, but as trajectories grow longer and more interference-heavy, \methodname{} more reliably preserves the evidence needed for downstream reasoning.

\begin{table}[t]
\centering
\small
\begin{tabular}{lccc}
\toprule
Method & S Recall & M Recall & L Recall \\
\midrule
Qwen-8B & 0.538 & 0.224 & 0.174 \\
text-embedding-3-small & 0.544 & 0.204 & 0.193 \\
text-embedding-3-large & 0.515 & 0.196 & 0.174 \\
RAPTOR & \textbf{0.694} & 0.263 & 0.206 \\
GraphRAG & 0.621 & 0.221 & 0.186 \\
HippoRAG 2 & 0.612 & 0.247 & 0.224 \\
\methodname{} & 0.416 & \textbf{0.387} & \textbf{0.292} \\
\bottomrule
\end{tabular}
\caption{Retrieval recall on CAME-Bench S/M/L. We report recall only for methods whose retrieved memory units can be mapped back to raw trajectory turns. Because many questions admit multiple valid evidence sets, this analysis is intended as a transparency-oriented diagnostic rather than the primary evaluation criterion.}
\label{tab:retrieval_recall}
\end{table}

%% file: sections/appendix/traj_number.tex
\section{Score Stability Across Independent Trajectories in \benchmarkname}

Because CAME-Bench contains a small number of long, high-density trajectories, we additionally assess how sensitive performance is to trajectory initialization. Concretely, we analyze pairwise F1 score differences across three independently generated trajectories (0, 1, and 2), summarize variance magnitude using Mean Absolute Deviation (MAD), and test for systematic bias using paired t-tests. We report this analysis for the Small and Medium partitions, which contain sufficient independent trajectories for a meaningful comparison.

Table~\ref{tab:init_stability} shows that initialization sensitivity diminishes as task complexity increases. In the Medium partition, no question type exhibits significant systematic bias ($p>0.05$), suggesting that performance is not driven by a single favorable trajectory seed. By contrast, the shorter Small partition shows mild sensitivity for Question Types 1 and 4, which involve incremental state revision and information synthesis. Across both scales, Question Type 3 remains highly stable, and Question Type 2 shows no significant bias despite moderate variance in the Medium setting. Overall, these results do not eliminate all finite-benchmark concerns, but they indicate that the observed gains are not an artifact of a particular initialization and become more stable in longer, more information-rich settings.

\begin{table}[t]
\centering
\footnotesize
\setlength{\tabcolsep}{4pt}
\begin{tabular}{@{}lccc@{}}
\toprule
Scale & QType & MAD & Bias $p$ \\
\midrule
\multirow{4}{*}{Small}
  & Q1 & 0.124 & 0.005 \\
  & Q2 & 0.000 & 0.108 \\
  & Q3 & 0.036 & 0.051 \\
  & Q4 & 0.267 & 0.020 \\
\midrule
\multirow{4}{*}{Medium}
  & Q1 & 0.243 & 0.756 \\
  & Q2 & 0.215 & 0.605 \\
  & Q3 & 0.023 & 0.977 \\
  & Q4 & 0.316 & 0.917 \\
\bottomrule
\end{tabular}
\caption{Sensitivity to trajectory initialization across independently generated trajectories. MAD denotes the mean absolute deviation of pairwise F1 differences across trajectory seeds; Bias $p$ reports the paired t-test for systematic differences.}
\label{tab:init_stability}
\end{table}

%% file: sections/appendix/method_prompt.tex
\subsection{Prompts Used by \methodname}
\label{app:stitch-prompts}

The following prompts are utilized by \methodname{} during annotation, inference, and retrieval-time processing. These prompts operate over existing dialogue trajectories and are executed online or during evaluation. They are instruction-only, contain no in-context examples, and are designed to generalize across domains.

We detail the prompts for thematic scope generation (\refsec{sec:thematic_scope}) in Listing~\ref{lst:context_scope_generation} and Listing~\ref{lst:scope_summary}; prompts for event labeling (\refsec{sec:event_label}) in Listing~\ref{lst:event_labeling}; prompts for key entity type labeling (\refsec{sec:entity_label}) in Listing~\ref{lst:entity_labeling}; and the prompt for memory snippet construction and canonical summary generation in Listing~\ref{lst:memory_snippet_generation}.

For retrieval, we utilize an LLM to select candidate labels for each component of contextual intent: thematic scope labels (Listing~\ref{lst:thematic_scope_selection}), event type labels (Listing~\ref{lst:event_label_selection}), and key entity types (Listing~\ref{lst:key_entity_type_selection}).

%% file: sections/appendix/evaluation_prompt.tex
\subsection{Prompts for Answer Generation and Evaluation}
\label{app:prompt_answer_gen_eval}

To ensure fair comparison, we use consistent prompts and models for answer generation. Listing~\ref{lst:response_generation_travel} details the answer generation prompts for the open-domain tasks in the LongMemEval benchmark and the travel planning subset of \benchmarkname. Listing~\ref{lst:response_generation_debate} provides the prompt used for answer generation in the debate domain of \benchmarkname.

To evaluate model performance across different question types, we employ two distinct evaluation protocols. For questions requiring a single, definitive answer (e.g., \textit{``What restaurant was selected for Day 1 dinner?''}), we utilize the prompt in Listing~\ref{lst:answer_eval_single}. This module assigns a binary label (CORRECT/WRONG) by verifying if the ground truth is present in the generated response. For complex queries involving multiple valid items (e.g., \textit{``What attractions were visited on Day 1?''}), we use the prompt in Listing~\ref{lst:answer_eval_set}. This module counts the number of valid candidates retrieved, enabling the calculation of fine-grained metrics such as Precision, Recall, and F1-score.

%% file: sections/appendix/trajectory_example.tex
\section{Benchmark Trajectory Showcase}
\label{app:travel_trajectory}

\cref{traj:travel-example} is a visualization of a travel planning trajectory from \benchmarkname. The interaction involves a \textbf{User} and a \textbf{Travel Agent} collaborating to plan a detailed multi-day itinerary. The dialogue encompasses constraint specification, information seeking (e.g., checking prices and ratings), option comparison, and final scheduling decisions. To prevent data contamination in future LLM evaluation, this example is slightly different from trajectories in the \benchmarkname.

\section{Debate Trajectory Showcase}
\label{app:debate_trajectory}

\cref{traj:debate-example} is a visualization of a debate trajectory from \benchmarkname. The interaction alternates between a \textbf{Pro-Side Debater} and a \textbf{Con-Side Debater} addressing the policy resolution: ``Resolved: The United States federal government should substantially reduce its restrictions on legal immigration to the United States.''\footnote{Topic sourced from https://www.speechanddebate.org.} The dialogue encompasses contentions, rebuttals, and summaries. To prevent data contamination in future LLM evaluation, this example utilizes a topic distinct from those contained in the actual benchmark dataset.

%% file: examples/benchmark_trajectory_travel_example.tex
\begin{trajectorybox}[traj:travel-example]{Condensed and truncated example of a travel planning trajectory in \benchmarkname}

\turn{\userbg{User}}{2025-05-15T08:00}{Let's focus on planning the itinerary for day 1. I'm eager to see what options we have and start organizing the activities.}

\turn{\agentbg{Travel Agent}}{2025-05-15T08:00}{Great, focusing on day 1 is a smart way to start shaping your trip! Let’s explore the best activities and sights to set a wonderful tone for your adventure. Whether you’re interested in cultural landmarks, nature spots, or vibrant local neighborhoods, we can tailor the day to your preferences. Just let me know what excites you most, and I’ll help organize an itinerary that balances excitement with ease, ensuring you have time to enjoy each experience without feeling rushed.}

\turn{\agentbg{Travel Agent}}{2025-05-15T08:00}{Starting your trip with a well-planned day can really enhance the overall flow and make sure you hit all the highlights comfortably. What kind of experiences are you most looking forward to on day 1?}

\turn{\userbg{User}}{2025-05-15T08:01}{One idea is the Daphne Laurel Hotel. It seems cozy and well-located, which is perfect since I prefer places that are easy to get around from. Plus, I usually appreciate a quiet atmosphere for resting after a day of sightseeing.}

\turn{\agentbg{Travel Agent}}{2025-05-15T08:01}{It is a truly charming choice, especially for travelers who value a quiet, unpretentious atmosphere combined with easy access to city attractions. Nestled in the serene old riverside quarter, this hotel is wrapped in a rich local story that adds a unique layer of character to your stay. Named after a gentle botanist who cared for the laurel groves nearby, the hotel’s air is still gently scented with these fragrant plants, which creates a peaceful ambiance right from the moment you arrive. What makes the Daphne Laurel particularly special is its blend of simplicity and warmth.}

\turn{\agentbg{Travel Agent}}{2025-05-15T08:01}{Its weathered stone walls and modest design make it feel more like a cozy retreat than a bustling hotel. Guests often speak fondly of the “kettle song,” a quaint tradition where the night porter hums softly while preparing communal tea in the lounge at dusk. This little ritual invites guests to relax and connect with a slower, more restful rhythm after a busy day of exploring the city. The rooms are practical yet thoughtful, many oriented to welcome the soft, warm morning light through delicate lace curtains, making waking up a gentle experience.}

\turn{\agentbg{Travel Agent}}{2025-05-15T08:01}{Free Wi-Fi keeps you connected without overwhelming the tranquil setting, and the continental breakfast provides a simple, satisfying start to your day. Located just steps from lantern-lit streets and lively markets, the hotel offers the perfect balance—close enough to the action but far enough to escape into calm. Many travelers find themselves extending their stay here, appreciating the steady, soothing pace and the blend of local history with genuine hospitality. If you’re someone who enjoys stepping into a space that feels both authentic and quietly restorative, It seems like an ideal fit for your trip.}

\turn{\userbg{User}}{2025-05-15T08:02}{One idea is to start the day at Hyperion Horizon Dining for breakfast. I’ve heard they have a great atmosphere and a variety of options, which might be perfect to fuel up for our day ahead.}

\turn{\agentbg{Travel Agent}}{2025-05-15T08:02}{It is a wonderful choice to start your day! This charming breakfast spot has a rich history, beginning as a cozy little nook on Lantern Lane that welcomed both early risers and late travelers seeking a comforting meal beneath the warm glow of oil lamps. The name itself conjures images of dawn breaking over distant peaks, symbolizing warmth and renewal—a perfect vibe to kick off your morning. What makes Hyperion Horizon truly special is its simple yet soul-soothing fare.}

\turn{\agentbg{Travel Agent}}{2025-05-15T08:02}{Their sun-kissed stew and earthy root flatbread are standout dishes, rooted in a story about a wandering cook who crafted these hearty recipes from foraged ingredients at daybreak. Guests often partake in the quaint tradition of leaving their spoons upright in their bowls, signaling shared experiences and quiet camaraderie over cups of fragrant spiced tea. The atmosphere is unpretentious and welcoming, with regulars softly chatting at the counter, and the soothing clatter of utensils blending into the hum of morning conversation. The dining room comes alive as the early light filters through weathered windows, mingling with the inviting aromas of freshly brewed herbal infusions.}

\turn{\agentbg{Travel Agent}}{2025-05-15T08:02}{Many visitors find themselves lingering here, not for extravagance, but for the comforting simplicity and warmth that wrap around you like a gentle embrace. It’s a spot that captures the steady rhythm of simple pleasures, setting a peaceful tone for your day ahead. Whether you’re looking to recharge or just enjoy a moment of calm, It offers an inviting and memorable start to your adventure.}

\turn{\userbg{User}}{2025-05-15T08:03}{One idea for breakfast is the Olympus Summit Dining Hall. It sounds like a great spot to start the day, especially if you like a lively atmosphere with a variety of options. Do you think it would be a good fit for us?}

\turn{\agentbg{Travel Agent}}{2025-05-15T08:03}{It is indeed a fantastic choice for breakfast, especially if you're looking to start the day with a unique and memorable experience. Perched on a gentle rise with a scenic view of the old riverside quarter, this dining hall is steeped in local lore, offering a setting that feels almost otherworldly. The name itself pays homage to a long-forgotten mountain peak and the legendary feast inspired by it, which sets the tone for an elevated meal. The house specialty, slow-roasted herb lamb with a honeyed glaze, is not only a dish but a ritual—symbolizing abundance and care.}

\turn{\agentbg{Travel Agent}}{2025-05-15T08:03}{This is a place where the meal unfolds at a thoughtful pace, inviting guests to savor each bite amidst the warm glow of lantern light and polished wood. While it is especially vibrant on weekend evenings, its serene elegance makes it a wonderful spot for a morning meal as well, providing a peaceful yet lively atmosphere. Many guests appreciate the attentive service that makes the dining experience feel personalized and special. If you enjoy a lively atmosphere paired with a variety of rich flavors and a touch of legend and tradition, It could be just the right start to your day.}

\turn{\userbg{User}}{2025-05-15T08:04}{One idea for dinner is Bellerophon Pegasus Dining. It sounds like a unique spot, and I usually appreciate places with a distinctive ambiance for evening meals. Maybe it could offer something memorable for our first night’s dinner?}

\turn{\agentbg{Travel Agent}}{2025-05-15T08:04}{It sounds like an enchanting choice for your first night's dinner! This restaurant is truly a gem nestled in the old quarters, where the charm of myth meets culinary artistry. The ambiance is uniquely refined, with softly glowing lanterns and crystal chandeliers that create a warm, inviting atmosphere perfect for an intimate evening. It's the kind of place where each dish is a story, inspired by the legendary hero who tamed Pegasus, and the signature "Skyward Feast" is a must-try—herb-encrusted meats paired with airy grains that evoke the sensation of flight.}

\turn{\agentbg{Travel Agent}}{2025-05-15T08:04}{The attention to detail here is remarkable, from the polished service to the ritual of waiting for the silver bell before dessert, making the dining experience feel both special and memorable. Many guests appreciate the slow, unhurried pace, allowing the meal to unfold as a leisurely, almost poetic journey. If you enjoy settings that blend elegance with a touch of mystery and history, this spot is likely to leave a lasting impression and set a wonderful tone for the rest of your trip.}

\turn{\userbg{User}}{2025-05-15T08:05}{One idea for lunch could be Ismene Courtyard Dining; it seems like a lovely spot with a nice ambiance, and I usually appreciate places that offer a relaxing courtyard setting for a midday meal.}

\turn{\agentbg{Travel Agent}}{2025-05-15T08:05}{It sounds like an absolutely enchanting choice for lunch, especially if you’re looking to enjoy a serene and elegant atmosphere. This dining spot is known for its exquisite blend of culinary artistry and storytelling, all set beneath a canopy of fragrant wisteria blossoms in a charming courtyard. The restaurant draws inspiration from the myth of Ismene, a muse of generosity and grace, which perfectly reflects in both the dishes served and the warm hospitality extended to guests. One of the highlights is their signature dish—a masterful creation featuring slow-roasted herbs and rare sea truffle, a recipe born from a legend about humble fishermen sharing their bounty with a wandering oracle.}

\turn{\agentbg{Travel Agent}}{2025-05-15T08:05}{This story sets the tone for the dining experience, weaving a sense of tradition and myth that elevates the meal beyond mere sustenance into something truly memorable. The staff’s white-glove service enhances the experience with a poised and attentive manner that ensures you can fully immerse yourself in the flavors without feeling rushed. As you arrive, especially around twilight, the courtyard transforms into a magical space with lanterns softly glowing and candlelight casting a warm, intimate ambiance. This creates a perfect setting not only for enjoying the meal but also for relaxing and savoring the moment, making it ideal for a peaceful midday break or even a lingering early evening meal.}

\turn{\agentbg{Travel Agent}}{2025-05-15T08:05}{The restaurant has a unique tradition where the dessert is served only when a soft bell chimes from the kitchen, signaling that the evening’s story is unfolding perfectly—adding an element of anticipation and ritual to the dining journey. Many guests find themselves drawn to stay longer, captivated by the combination of exquisite cuisine, elegant surroundings, and the inviting atmosphere that encourages leisurely enjoyment and convivial conversation. It’s a place that feels both intimate and expansive, where each course is part of a carefully choreographed experience. This makes It particularly well-suited for those who appreciate a relaxed yet refined dining pace, where the meal itself is as much about savoring time and stories as it is about the food.}

\turn{\agentbg{Travel Agent}}{2025-05-15T08:05}{Whether you’re seeking a sophisticated spot for a quiet lunch or a charming place to unwind in the afternoon, It promises a memorable experience that gently elevates the everyday into the extraordinary. It’s a choice that not only pleases the palate but also nourishes the spirit with its unique blend of history, ambiance, and culinary excellence.}

\turn{\userbg{User}}{2025-05-15T08:06}{Could you tell me about the price range for breakfast at the 1th restaurant I raised before as the breakfast of Day 1's plan? I want to make sure it fits within our budget before we decide.}

\turn{\agentbg{Travel Agent}}{2025-05-15T08:06}{Certainly! Breakfast at It is priced at around \$76.22, which places it in the higher price range for breakfast options. This price reflects not only the quality of ingredients used but also the exceptional dining experience offered. The restaurant boasts a solid reputation, supported by positive guest reviews that highlight its refined ambiance and attentive service.}

\turn{\agentbg{Travel Agent}}{2025-05-15T08:06}{Many guests appreciate the thoughtfully curated breakfast menu, which features a blend of classic favorites and innovative dishes prepared with a gourmet touch. It's an excellent choice if you're looking to start your day with a luxurious meal in an elegant setting. Keep in mind that this upscale dining spot is well-suited for travelers who enjoy a leisurely, indulgent breakfast experience, possibly to celebrate a special occasion or to simply treat themselves. If you're mindful of your budget, this is a premium choice, but one that many find worthwhile for the quality and atmosphere it provides.}

\turn{\agentbg{Travel Agent}}{2025-05-15T08:06}{If you prefer a more casual or budget-friendly option, there might be other dining venues to consider, but It stands out for those who value both taste and presentation. Would you like more information on what the breakfast menu typically includes or suggestions for nearby alternatives?}

\turn{\userbg{User}}{2025-05-15T08:07}{Could you tell me more about the price range and the ratings for breakfast at the 2th restaurant I raised before as the breakfast of Day 1's plan? I’m trying to get a sense of whether it’s a good value and if it’s well-liked by other travelers.}

\turn{\agentbg{Travel Agent}}{2025-05-15T08:07}{It offers breakfast at around \$170.24, placing it in the higher price range for morning meals, which suggests it caters to guests looking for a premium or unique dining experience. However, its rating is 2.7, which indicates mixed feedback from previous guests. This score might reflect some inconsistency in service or food quality, or perhaps it doesn't quite meet the expectations set by its pricing. Breakfast here might feature specialty dishes or an exclusive ambiance, likely situated in a scenic or prestigious location given the 'Summit' in its name, appealing to travelers seeking a memorable start to their day.}

\turn{\agentbg{Travel Agent}}{2025-05-15T08:07}{While it may not be the best value for all guests, some visitors might appreciate the atmosphere or the quality of ingredients despite the higher cost. It might suit those who prioritize the experience and are less sensitive to price, but if you're looking for a budget-friendly or consistently highly rated breakfast spot, this may require some consideration. Keep in mind it's always good to read recent reviews or ask about specific menu options to ensure it aligns with your preferences.}

\turn{\userbg{User}}{2025-05-15T08:08}{Let's learn more about the 1st activity we want to schedule during the day we are discussing. One idea is to visit the Andromeda Galaxy Dome. I think it could be a fascinating spot since I love astronomy, and it seems like a peaceful place to unwind and learn something new.}

\turn{\agentbg{Travel Agent}}{2025-05-15T08:08}{It sounds like a wonderful choice, especially for someone with a passion for astronomy. Nestled at the edge of the old riverside quarter, this attraction offers a serene escape into the cosmos. Its geodesic structure is a contemporary homage to ancient stargazing, creating a unique atmosphere that connects visitors with the night sky in a meaningful way. The dome was created from the vision of a modest stargazer who wanted to share the wonders of the universe affordably, and today it stands as an inviting and economical destination that welcomes both solo explorers and families.}

\turn{\agentbg{Travel Agent}}{2025-05-15T08:08}{One of the highlights of the dome is the softly glowing projection of the Andromeda constellation, which is accompanied by a gentle musical score during twilight hours. This creates a mesmerizing and calming experience that draws locals and visitors who appreciate the peaceful ambiance before the evening market buzz begins nearby. Guests usually spend about an hour here, seated on simple benches arranged around the dome’s edge, where they can quietly reflect or share whispered legends, such as the story of a hidden star that is said to bring clarity of purpose if spotted just before the dome’s quiet hour. The setting is thoughtfully designed to maintain a tranquil environment; for instance, there are gentle stairs for easy access, and pets are kindly asked to stay outside to preserve the peacefulness.}

\turn{\agentbg{Travel Agent}}{2025-05-15T08:08}{The dome’s charm lies in its humble yet evocative tribute to the vast universe—it's a space that feels both intimate and mysterious, encouraging visitors to lose themselves in the wonder of the stars. This spot is ideal if you’re looking for a place to unwind and soak in some cosmic inspiration. Many guests appreciate the dome’s blend of quiet reflection and gentle storytelling, which makes it more than just an astronomy attraction but also a meaningful cultural experience. If you enjoy calm, educational environments with a touch of magic, It will surely resonate with you.}

\turn{\agentbg{Travel Agent}}{2025-05-15T08:08}{It’s also worth noting that its location near the old riverside quarter means you could easily combine this visit with a stroll through charming historic streets afterward. Overall, it’s a peaceful, accessible, and enriching destination for your itinerary.}

\turn{\userbg{User}}{2025-05-15T08:09}{Let's learn more about the 1st activity we want to schedule during the day we are discussing. Could you tell me about the ticket prices for the the first place I raised before as the first activity of the Day 1 's plan? I'm trying to figure out if it fits within our budget for the day.}

\turn{\agentbg{Travel Agent}}{2025-05-15T08:09}{The ticket price for It is \$33.94, which positions it in the mid-range category for attractions of its kind. This price reflects a balanced investment for an immersive and engaging experience, often featuring high-tech visual displays and interactive exhibits that bring the cosmos to life. The Dome tends to be well-rated for its captivating presentations and educational value, making it a popular choice among visitors interested in astronomy and science. Many guests appreciate the blend of entertainment and learning it offers, making it suitable for families, couples, and solo travelers alike.}

\turn{\agentbg{Travel Agent}}{2025-05-15T08:09}{It’s worth considering that this price includes access to specialized shows that aren’t commonly found at other venues, adding to the overall value. Additionally, the Dome is usually located within a science center or museum complex, often surrounded by amenities such as cafes and gift shops, which can enhance your visit. So, if your budget allows for a mid-tier experience that combines fun and education, It is an excellent pick for your day’s itinerary.}

\turn{\userbg{User}}{2025-05-15T08:10}{Let's learn more about the 1st activity we want to schedule during the day we are discussing. I wonder if we could try visiting the Nyx Twilight Observatory. It seems like a unique spot, especially for catching a stunning sunset and some star gazing later in the evening. Plus, I usually enjoy quiet, peaceful places like that.}

\turn{\agentbg{Travel Agent}}{2025-05-15T08:10}{It sounds like a truly enchanting choice for your visit, especially if you cherish quiet, reflective places that offer a deep connection to nature and the cosmos. Perched atop the mist-kissed cliffs of the old riverside quarter, this observatory is more than just a vantage point; it’s steeped in local lore and offers an almost mystical experience as twilight descends. Named after Nyx, the guardian of dusk, it embodies the magic of the transition between day and night, creating a setting that feels both timeless and intimate. Visitors often arrive just as daylight fades, drawn to the observatory’s distinctive domed silhouette standing out against the evening sky.}

\turn{\agentbg{Travel Agent}}{2025-05-15T08:10}{It’s a place where you can spend hours tracing constellations through carefully polished lenses, guided by the quiet, knowledgeable night porter who enhances the experience with a calm and welcoming presence. The atmosphere is enriched by the subtle scents of aged wood and cooled stone, transporting guests to a serene, almost otherworldly state. One of the most cherished experiences here is the silent “twilight walk” along a lantern-lined path, a ritual said to draw the stars nearer for a clearer view. It’s a perfect moment to savor peace and quiet before settling in for the celestial show.}

\begin{center}
    \vspace{1em}
    \textit{... [Trajectory truncated. Total turns: 186. Showing first 40.] ...}
\end{center}

\label{app:travel_trajectory_example}
\end{trajectorybox}

%% file: examples/benchmark_trajectory_debate_example.tex
\begin{trajectorybox}[traj:debate-example]{Condensed and truncated example of a debate trajectory in \benchmarkname}
% --- Turn 1: Pro-Side Proposal ---
\debateturn{\prolabel{Pro-Side Debater}}{2025-01-01T00:00}{%
Reducing restrictions on legal immigration fills labor shortages, increases productivity, and stimulates economic growth—especially in healthcare, STEM, agriculture, and service sectors facing chronic worker gaps.

That contention captures the central practical case: the U.S. faces persistent, observable shortages in specific occupations and industries, and this is precisely what immigration policy can correct. Bringing more legal workers into those fields does three things at once: it supplies the needed hands and skills now, it raises overall output by filling bottlenecks (so capital and domestic labor are used more efficiently), and it lowers prices or stabilizes wages in affected services—benefiting consumers and businesses.

Policy design matters. This is not advocating unfocused open borders; it’s about targeted, evidence-driven liberalization: raise employment-based caps, create sector-specific visas tied to demonstrated shortages, and expedite permanent residency for occupations with sustained demand. These measures direct immigration where it produces the most economic benefit while guarding labor standards.
}

% --- Turn 6: Con-Side Proposal ---
\debateturn{\conlabel{Con-Side Debater}}{2025-01-01T00:15}{%
Contention: Expanded legal immigration could increase competition for jobs, particularly for low-skilled or vulnerable workers, putting downward pressure on wages and employment opportunities.

This contention matters because it highlights a predictable economic mechanism and a real distributional effect. When that many more workers enter a labor market, especially without matching increases in demand for the specific types of labor they supply, this raises labor supply relative to demand in certain occupations and places. That, in turn, tends to reduce the bargaining power of incumbent low-wage workers and can compress their wages or raise unemployment.

There is empirical backing for these concerns. Several respected studies find that immigrant inflows can impose downward pressure on wages and employment for earlier arrivals and for native workers in the same low-skill segments. Rather than sweeping reductions, sensible alternatives include skill- and sector-targeted admissions and stronger wage and labor-enforcement provisions.
}

% --- Turn 11: Con-Side Defense ---
\debateturn{\conlabel{Con-Side Debater}}{2025-01-01T00:30}{%
Here’s why that still stands.

First, timing and funding structure matter. Many critical services are financed and planned on multi-year cycles. When population rises quickly, local governments cannot instantly expand capacity: hiring qualified staff and building classrooms take time.

Second, concentrated impacts produce hotspots. Immigrants are not evenly dispersed. They tend to settle where jobs and social networks already exist, which concentrates demand in particular school districts and neighborhoods. One overwhelmed county or city can experience overcrowded classrooms and longer emergency-room waits that aren’t visible in aggregate statistics.

All of this shows the contention is not just about aggregate numbers of people but about the practical, institutional realities of local service delivery.
}

% --- Turn 13: Con-Side Attack ---
\debateturn{\conlabel{Con-Side Debater}}{2025-01-01T00:36}{%
I disagree with that contention — it overstates a simple policy link and ignores important trade-offs and practical limits.

1) Correlation isn’t causation. Those high-profile innovators often succeed because of U.S. universities and venture capital networks. Easing all restrictions won’t reproduce those specific institutional supports.

2) Measurement problems bias the claim. Counting patents or startup founders inflates one dimension of economic contribution but ignores quality and survivorship bias.

3) Distributional and fiscal consequences are overlooked. Large, rapid inflows can raise housing costs and strain schools, harms that fall on lower-income native workers.

In short, the contention assumes a direct, unambiguous causal effect from broad liberalization to more innovation. That link is weak.
}

% --- Turn 52: Pro-Side Summary ---
\debateturn{\prolabel{Pro-Side Debater}}{2025-01-01T02:33}{%
This case argues that substantially reducing federal restrictions on legal immigration produces clear economic, fiscal, and social benefits. Those changes would fill chronic labor shortages and raise productivity in high-need fields—healthcare, STEM, agriculture, and services—so that the economy can grow and operate efficiently. This shift would also make it easier to attract and retain global talent who disproportionately found startups and drive research. Finally, this approach offsets population aging and decline, revitalizes communities, and reaffirms America’s identity as a nation of opportunity and pluralism.
}

% --- Turn 53: Con-Side Summary ---
\debateturn{\conlabel{Con-Side Debater}}{2025-01-01T02:36}{%
These are the core reasons we oppose substantially reducing federal immigration restrictions. First, this change would intensify competition for jobs, especially among low-skilled and vulnerable workers. Second, this policy would likely accelerate brain drain from poorer countries. Third, these inflows would place real pressure on local public services that often cannot scale funding quickly enough. Finally, a rapid expansion risks overwhelming integration systems, producing social tension and political backlash. Together, these points form a coherent argument that reducing restrictions carries significant costs that should not be underestimated.
}

\begin{center}
    \vspace{1em}
    \textit{... [Trajectory condensed for showcase. Full benchmark contains 53 detailed turns.] ...}
\end{center}

\label{app:debate_trajectory_example}
\end{trajectorybox}

%% file: prompts/context_scope_generation.tex
\begin{prompt}
class TurnScopeSignature(dspy.Signature):
    """
    Determine which context scope this utterance belongs to.

    Guidelines
    1. The granularity of the context scope should be similar to as previous context scopes.
    2. Combined with the utterance, you must consider the best context scope to quickly partition the conversation into different scopes.
    3. Observe the current utterance carefully to identify whether it signals a context scope transition or continuation within the same context scope.
    4. Compare with prior_structured_notes to check if the utterance refers back to or continues a prior context scope.
    5. Default to continuity:
       Read full utterance carefully, if the utterance does not explicitly introduce a new context scope, assign the same context scope as the most recent relevant prior note.
    6. Detect transitions:
       When the speaker introduces a transition to a new context, assign a new context scope label to reflect that shift.
    7. Maintain consistency:
       ALWAYS check the existing_context_scopes list first. If the utterance refers to a topic that matches an existing scope (even if semantically similar), reuse that exact string form. Only create a new scope if no existing scope matches.
    """

    turn_id: str = dspy.InputField()
    role: str = dspy.InputField()
    utterance: str = dspy.InputField()
    conversation_type: str  = dspy.InputField()
    prior_structured_notes: str = dspy.InputField(
        description="Previously predicted scopes with turn_id, role, and context_scope (last 20 turns)."
    )
    existing_context_scopes: str = dspy.InputField(
        description="JSON list of all unique context scope labels that have been used in this conversation so far. Check this list first to reuse existing scopes before creating new ones."
    )
    context_scope: str = dspy.OutputField()
\end{prompt}

%% file: prompts/canonical_summary_generation.tex
\begin{prompt}
class ContextNoteKeepingSignature(dspy.Signature):
    """Your goal is to maintain a clear, scope-explicit summaries.

    Rules:
        1. Resolve ambiguity or confusion

    - The prior context notes are from the same scope as the current segment.
    - The prior context notes provide all the information mentioned in the same scope
    - Use prior context notes safely to resolve all vague references to disambiguate the current segment

    2. Content

    - Capture only new, semantically meaningful developments
    - List important new targets within each scope as numbered or bulleted items for clarity
    - You should not list any vague references or pronouns that are not resolved by the prior context notes
    """

    context_scope: str = dspy.InputField(description="Name of the context scope shared by this segment.")
    prior_segment_level_notes: str = dspy.InputField(description="JSON list of previous notes for this same scope, ordered chronologically.")
    segment_turns: str = dspy.InputField(description="JSON list of dictionaries with keys (turn_index, role, utterance) for each turn in this scope segment.")
    segment_level_note: str = dspy.OutputField(description="1-3 sentence update capturing new developments for this scope.")
    
\end{prompt}

%% file: prompts/event_type_generation.tex
\begin{prompt}
class EventTypeLabelGenerationSignature(dspy.Signature):

    """
    Generate a list of event type labels that describe the different types of discussions or events that occur in a conversation. Given the first 50 turns of a conversation and the dataset type, identify the distinct event types or discussion themes present. Event types should be concise labels that capture the nature of the events and can be used to describe multiple similar events.
    """

    first_turns_summary: str = dspy.InputField(description="Summary of the first 50 turns (first sentence + '...' + last sentence of each turn)")

    dataset_type: str = dspy.InputField(description="Type of dataset")

    event_type_labels: list[str] = dspy.OutputField(description="List of event type labels for this conversation")

class EventTypeLabelValidationSignature(dspy.Signature):

    """
    Given a batch of consecutive turns and the current set of event type labels, determine if new event types are present that are not covered by existing labels.  Also, check if the existing labels have appropriate coverage and description to describe the batch of turns. Make least number of changes to the existing labels. Only add new labels if necessary.  Return ONLY the new event type labels that should be added to the existing set.

    If the existing labels are sufficient, return an empty list. Do NOT return the full updated list, only return the new labels to add.
    """

    batch_turns_summary: str = dspy.InputField(description="Summary of a batch of 50 consecutive turns (first sentence + '...' + last sentence of each turn)")

    existing_event_labels: list[str] = dspy.InputField(description="List of existing event type labels")

    dataset_type: str = dspy.InputField(description="Type of dataset")

    new_event_labels: list[str] = dspy.OutputField(description="ONLY the new event type labels to ADD to the existing set. Return empty list if no new labels needed.")
    
\end{prompt}

%% file: prompts/key_entity_type_generation.tex
\begin{prompt}
class FineGrainedFunctionalSeedSignature(dspy.Signature):

    """You are given (1) a dataset description and (2) a sample of turns. Your task is to derive a list of fine-grained functional type that describe how specific details, facts, references, options, or objects are being used to advance the task in this dataset. IMPORTANT: - You ARE generating a list of pragmatic, task-driven functional type names that that represent how fine-grained details participate in argumentation, reasoning, decision-making, planning, critique, or exploration. - Think in terms of what work each detail is doing for the speaker or agent. Method: 1. Understand the dataset's task type from the description. 2. From the sample turns, identify recurring uses of specific details: - What are these details achieving pragmatically? - How do they drive the conversation or task forward? 3. Abstract the observed uses into general *functional types*. - Names should be concise noun or noun-phrase labels (2-4 words). - Different names should distinguish from each other. - The number of functional type names is limited, so you should use the most general and neutral names that can cover meaningful details in the dataset. 4. Order the list from most representative to least representative.
    """

    dataset_description: str = dspy.InputField()

    sample_turns: str = dspy.InputField()

    functional_type_seeds: list[str] = dspy.OutputField()


\end{prompt}

%% file: prompts/memory_snippet.tex
\begin{prompt}
class TurnNoteContentSignature(dspy.Signature):
    """
    Generate structured note content (act, target, and note_text) for a turn, conditioned on the known context_scope and event types.

    Objective:
    Express the speaker's communicative intent in a concise, structured way, identifying their pragmatic action (act), the entity/topic it concerns (target), and a short natural-language summary (note_text).

    Guidelines

    1. Act identification:
       Based on the dataset type and the role of the speaker, determine the speaker's pragmatic act
    2. Target identification:
       Identify the specific entity, topic, claim, or object that drives the discussion.

    - If a concrete object or entity name is explicitly mentioned and drives the discussion, select that as the target.
    - If not explicitly mentioned, infer the implicit object from the semantic meaning of the utterance.
    - When ambiguous, refer to prior structured_notes within the same context scope or event types to infer or resolve the referent.
    - Resolve pronouns or elliptical expressions (e.g., "it," "this one," "there") by tracing to previously mentioned entities or items.

    Note text composition:
    Write one short, functional sentence summarizing what the speaker is doing and about what.

    - If a concrete object or entity name is explicitly mentioned and drives the discussion, include the entity name described accurately in the note text.
    - Focus on communicative intent and salient target attributes.
    - Do not include irrelevant details or paraphrase entire utterances.
    4. Functional types selection:
       - The provided functional type candidates are a list of pragmatic and task-driven high-level types aggregating the functions of meaningful details in the dataset.
       - Read the utterance carefully and select 0 to any number of functional types that cover the details that drive the utterance.
    5. Context-awareness:
       - Ensure the generated act and target align with the given context_scope.
       - The scope defines what sub-topic or thread this turn contributes to.
         Use segment_level_notes to recall prior developments within this scope.
    6. Event-type conditioning:
       - Use event_types to refine your interpretation
    7. Consistency check:
       - If multiple prior turns have similar acts or targets under the same scope, maintain consistent terminology and phrasing.
    """

    turn_id: str = dspy.InputField()
    dataset_type: str = dspy.InputField()
    role: str = dspy.InputField()
    utterance: str = dspy.InputField()
    context_scope: str = dspy.InputField()
    event_types: str = dspy.InputField(description="Comma-separated event type labels for the current turn")
    prior_structured_notes: str = dspy.InputField(description="Prior notes sharing the same context scope or event types, used for disambiguation.")
    segment_level_notes: str = dspy.InputField(description="JSON list of segment-level summaries for this scope observed up to the current turn.")
    functional_type_seeds_candidates: list[str] = dspy.InputField(description="List of functional type candidates to choose from for the turn.")
    act: str = dspy.OutputField()
    target: str = dspy.OutputField()
    note_text: str = dspy.OutputField()
    functional_type_seeds: list[str] = dspy.OutputField()
\end{prompt}

%% file: prompts/context_scope_selection.tex
\begin{prompt}
class ContextScopeLabelSelectionSignature(dspy.Signature):

    """Select appropriate number of labels for context_scope field to filter conversation turns that can help answer the question.

    Your task is, read the label names to judge if the label is useful to answer the question. Don't overthink. The useful label names will implicitly or explicitly tell relevant information to the question. Then select an appropriate number(can be 0) of useful labels.

    Select the label names that are likely to be useful to answer the question.
    """

    question: str = dspy.InputField()

    context_scope_candidates: List[str] = dspy.InputField(description="All unique context_scope values in the dataset")

    selected_context_scopes: List[str] = dspy.OutputField(description="An appropriate number of selected context_scope values (empty list if none relevant)")

    reasoning: str = dspy.OutputField(description="Reasoning for the selected context_scope values")
\end{prompt}

%% file: prompts/event_type_selection.tex
\begin{prompt}
class EventTypeLabelSelectionSignature(dspy.Signature):

    """Select appropriate number of labels for event_types field to filter conversation turns that can help answer the question.

    Your task is, read the label names to judge if the label is useful to answer the question. Don't overthink. The useful label names will implicitly or explicitly tell relevant information to the question. Then select an appropriate number(can be 0) of useful labels.

    Select the label names that are likely to be useful to answer the question.
    """

    question: str = dspy.InputField()

    event_type_candidates: List[str] = dspy.InputField(description="All unique event_type values in the dataset")

    selected_event_types: List[str] = dspy.OutputField(description="An appropriate number of selected event_type values (empty list if none relevant)")

    reasoning: str = dspy.OutputField(description="Reasoning for the selected event_type values")
\end{prompt}

%% file: prompts/key_entity_type_selection.tex
\begin{prompt}
class FunctionalTypeSeedsLabelSelectionSignature(dspy.Signature):

    """Select appropriate number of labels for functional type seeds field to filter conversation turns that can help answer the question.

    Your task is, read the label names to judge if the label is useful to answer the question. Don't overthink. The useful label names will implicitly or explicitly tell relevant information to the question. Then select an appropriate number(can be 0) of useful labels.

    Select the label names that are likely to be useful to answer the question.
    """

    question: str = dspy.InputField()

    functional_type_seeds_candidates: List[str] = dspy.InputField(description="All unique functional type seeds values in the dataset")

    selected_functional_type_seeds: List[str] = dspy.OutputField(description="An appropriate number of selected functional type seeds values (empty list if none relevant)")

    reasoning: str = dspy.OutputField(description="Reasoning for the selected functional type seeds values")
\end{prompt}

%% file: prompts/response_generation_general_travel.tex
\begin{prompt}
class FreeformAnswerGenerator(dspy.Signature):
    """
    Answer a freeform question using the given set of retrieved conversation turns. 
    Read carefully the question. The retrieved turns are a set of relevant and useful turns to answer the question. You need review these turns carefully to answer the question correctly. 
    In case of doubt or uncertainty, do not guess, say "I don't know".
    """
    question: str = dspy.InputField(description="The question to be answered")
    task_setting: str = dspy.InputField(description="The setting of the task you are answering")
    retrieved_turns: str = dspy.InputField(
        description="A list of conversation turn contents relevant to the question.")
    answer_reasoning: str = dspy.OutputField(description="Provide one sentence reasoning on your answer")
    output: str = dspy.OutputField(
        description="An answer to the question. If the question is not answerable, return 'Question not answerable'")
\end{prompt}

%% file: prompts/response_generation_debate.tex
\begin{prompt}
class FreeformAnswerGenerator_Debate(dspy.Signature):
    """
    Answer a freeform question using the given set of retrieved conversation turns. 
    Read carefully the question. The retrieved turns are a set of relevant and useful turns to answer the question. You need review these turns carefully to answer the question correctly. You should not include any unrelevant information in your answer. Give the shortest answer that correctly answer the question.
    In case of doubt or uncertainty, do not guess, say "I don't know".
    """
    question: str = dspy.InputField(description="The question to be answered")
    task_setting: str = dspy.InputField(description="The setting of the task you are answering")
    retrieved_turns: str = dspy.InputField(
        description="A list of conversation turn contents relevant to the question.")
    answer_reasoning: str = dspy.OutputField(description="Provide one sentence reasoning on your answer")
    output: str = dspy.OutputField(
        description="An answer to the question. If the question is not answerable, return 'Question not answerable'")
\end{prompt}

%% file: prompts/answer_evaluator_single.tex
\begin{prompt}
class LLMAnswerEvaluatorSignature(dspy.Signature):
    """Your task is to label an answer to a question as \"CORRECT\" or \"WRONG\".
    You will be given the following data: (1) a question, (2) a gold answer, and (3) a generated answer.
    The gold answer is concise and contains the ground-truth information. The generated answer might be longer.
    Be generous: if the generated answer contains the gold answer information (even verbatim inside a longer response),
    mark it as CORRECT. Otherwise, mark it as WRONG. Respond with either CORRECT or WRONG, and provide a brief reasoning."""

    question: str = dspy.InputField(description="The question being evaluated")
    gold_answer: str = dspy.InputField(description="Gold answer text")
    generated_answer: str = dspy.InputField(description="Generated answer text")
    verdict: str = dspy.OutputField(description="CORRECT or WRONG")
    reasoning: str = dspy.OutputField(description="Short justification")
\end{prompt}

%% file: prompts/answer_evaluator_set.tex
\begin{prompt}
class LLMAnswerEvaluatorNumberSignature(dspy.Signature):
    """"Your task is to give the number of correct candidates to a question appearing in the generated answer.
    You will be given the following data: (1) a question, (2) a gold answer containing several correct candidates, and (3) a generated answer.
    The gold answer and contains the ground-truth candidates. Go through each candidate in the generated answer carefully and check if any of them appear in the gold answer list. 
    Respond with the number of correct candidates covered by the generated answer, and provide a brief reasoning."""

    question: str = dspy.InputField(description="The question being evaluated")
    gold_answer: List[str] = dspy.InputField(description="Gold answer candidates (list)")
    generated_answer: List[str] = dspy.InputField(description="Generated answer candidates (list)")
    number_of_correct_candidates: int = dspy.OutputField(description="Number of correct candidates")
    reasoning: str = dspy.OutputField(description="Short justification")


\end{prompt}